\documentclass[12pt]{iopart}

\usepackage{comment}
\usepackage{acronym}
\usepackage{todonotes}
\usepackage{subcaption}
\expandafter\let\csname equation*\endcsname\relax
\expandafter\let\csname endequation*\endcsname\relax
\usepackage{amsmath}
\usepackage{algorithm,algpseudocode,float}
\usepackage{placeins}
\usepackage[commandnameprefix=ifneeded]{changes}
\usepackage[T1]{fontenc}

\usepackage[
  backend=biber,
  style=ieee,   
  citestyle=numeric-comp,
  doi=false,
  isbn=false,
  url=false,
  date=year,
  maxbibnames=10,   
  maxcitenames=10,
  minnames=1    
]{biblatex}

\usepackage[
    colorlinks=true,
    linkcolor=black,
    citecolor=black,
    filecolor=black,
    urlcolor=black,
    plainpages=false,
    pdfpagelabels,
    breaklinks=true,
    pdfusetitle,
    ]{hyperref}

\usepackage{tikz}
\usepackage{circuitikz}

\usetikzlibrary{calc} 

\acrodefplural{RAM}[RAMs]{Random Access Memories}
\acrodefplural{RRAM}[R-RAMs]{Resistive Random Access Memories}
\acrodefplural{STT-MRAM}[STT-MRAMs]{Spin-Transfer Torque Magnetic Random Access Memories}
\acrodef{ADC}[ADC]{Analog to Digital Converter}
\acrodef{ADEX}[AdExp-I\&F]{Adaptive-Exponential Integrate and Fire}
\acrodef{AL}[AL]{Antennal Lobe}
\acrodef{AER}[AER]{Address-Event Representation}
\acrodef{AEX}[AEX]{AER EXtension board}
\acrodef{AE}[AE]{Address-Event}
\acrodef{AFM}[AFM]{Atomic Force Microscope}
\acrodef{AGC}[AGC]{Automatic Gain Control}
\acrodef{AMDA}[AMDA]{AER Motherboard with D/A converters}
\acrodef{ANN}[ANN]{Attractor Neural Network}
\acrodef{API}[API]{Application Programming Interface}
\acrodef{ARM}[ARM]{Advanced RISC Machine}
\acrodef{ASIC}[ASIC]{Application Specific Integrated Circuit}
\acrodef{BCM}[BMC]{Bienenstock-Cooper-Munro}
\acrodef{BD}[BD]{Bundled Data}
\acrodef{BEOL}[BEOL]{Back-end of Line}
\acrodef{BG}[BG]{Bias Generator}
\acrodef{BMI}[BMI]{Brain-Machince Interface}
\acrodef{CAD}[CAD]{Computer Aided Design}
\acrodef{CAM}[CAM]{Content Addressable Memory}
\acrodef{CAVIAR}[CAVIAR]{Convolution AER Vision Architecture for Real-Time}
\acrodef{CFC}[CFC]{Current to Frequency Converter}
\acrodef{CCN}[CCN]{Cooperative and Competitive Network}
\acrodef{CDR}[CDR]{Clock-Data Recovery}
\acrodef{CFC}[CFC]{Current to Frequency Converter}
\acrodef{CHP}[CHP]{Communicating Hardware Processes}
\acrodef{CNN}[CCN]{Convolutional Neural Network}
\acrodef{CMIM}[CMIM]{Metal-insulator-metal Capacitor}
\acrodef{CML}[CML]{Current Mode Logic}
\acrodef{CMOL}[CMOL]{``Hybrid CMOS nanoelectronic circuits''}
\acrodef{CMOS}[CMOS]{Complementary Metal-Oxide-Semiconductor}
\acrodef{CNN}[CCN]{Convolutional Neural Network}
\acrodef{COTS}[COTS]{Commercial Off-The-Shelf}
\acrodef{CPG}[CPG]{Central Pattern Generator}
\acrodef{CPLD}[CPLD]{Complex Programmable Logic Device}
\acrodef{CPU}[CPU]{Central Processing Unit}
\acrodef{CSP}[CSP]{Constraint Satisfaction Problem}
\acrodef{CV}[CV]{Coefficient of Variation}
\acrodef{DAC}[DAC]{Digital to Analog Converter}
\acrodef{DAS}[DAS]{Dynamic Auditory Sensor}
\acrodef{DAVIS}[DAVIS]{Dynamic and Active Pixel Vision Sensor}
\acrodef{DBN}[DBN]{Deep Belief Network}
\acrodef{DFA}[DFA]{Deterministic Finite Automaton}
\acrodef{DI}[DI]{delay insensitive}
\acrodef{DMA}[DMA]{Direct Memory Access}
\acrodef{DNF}[DNF]{Dynamic Neural Field}
\acrodef{DNN}[DNN]{Deep Neural Network}
\acrodef{DOF}[DOF]{Degrees of Freedom}
\acrodef{DPE}[DPE]{Dynamic Parameter Estimation}
\acrodef{DPI}[DPI]{Differential Pair Integrator}
\acrodef{DR-RZ}[DR-RZ]{Dual-Rail Return-to-Zero}
\acrodef{DRAM}[DRAM]{Dynamic Random Access Memory}
\acrodef{DR}[DR]{Dual Rail}
\acrodef{DSP}[DSP]{Digital Signal Processor}
\acrodef{DVS}[DVS]{Dynamic Vision Sensor}
\acrodef{EBL}[EBL]{Electron Beam Lithography}
\acrodef{EDVAC}[EDVAC]{Electronic Discrete Variable Automatic Computer}
\acrodef{EIN}[EIN]{Excitatory-Inhibitory Network}
\acrodef{EM}[EM]{Expectation Maximization}
\acrodef{EPSC}[EPSC]{Excitatory Post-Synaptic Current}
\acrodef{EPSP}[EPSP]{Excitatory Post-Synaptic Potential}
\acrodef{FDSOI}[FD-SOI]{Fully-Depleted Silicon on Insulator}
\acrodef{FET}[FET]{Field-Effect Transistor}
\acrodef{FFT}[FFT]{Fast Fourier Transform}
\acrodef{FI}[F-I]{Frequency-Current}
\acrodef{FPGA}[FPGA]{Field Programmable Gate Array}
\acrodef{FSA}[FSA]{Finite State Automaton}
\acrodef{FSM}[FSM]{Finite State Machine}
\acrodef{GOPS}[GOPS]{Giga-Operations per Second}
\acrodef{GPU}[GPU]{Graphical Processing Unit}
\acrodef{GUI}[GUI]{Graphical User Interface}
\acrodef{HAL}[HAL]{Hardware Abstraction Layer}
\acrodef{HH}[H\&H]{Hodgkin \& Huxley}
\acrodef{HMM}[HMM]{Hidden Markov Model}
\acrodef{HRS}[HRS]{High-Resistive State}
\acrodef{HR}[HR]{Human Readable}
\acrodef{HSE}[HSE]{Handshaking Expansion}
\acrodef{HW}[HW]{Hardware}
\acrodef{ICT}[ICT]{Information and Communication Technology}
\acrodef{IC}[IC]{Integrated Circuit}
\acrodef{IF2DWTA}[IF2DWTA]{Integrate \& Fire 2--Dimensional WTA}
\acrodef{IFSLWTA}[IFSLWTA]{Integrate \& Fire Stop Learning WTA}
\acrodef{IF}[I\&F]{Integrate-and-Fire}
\acrodef{IMU}[IMU]{Inertial Measurement Unit}
\acrodef{INCF}[INCF]{International Neuroinformatics Coordinating Facility}
\acrodef{INI}[INI]{Institute of Neuroinformatics}
\acrodef{IO}[I/O]{Input/Output}
\acrodef{IoT}[IoT]{Internet of Things}
\acrodef{IPSC}[IPSC]{Inhibitory Post-Synaptic Current}
\acrodef{IPSP}[IPSP]{Inhibitory Post-Synaptic Potential}
\acrodef{IP}[IP]{Intellectual Property}
\acrodef{ISI}[ISI]{Inter-Spike Interval}
\acrodef{IoT}[IoT]{Internet of Things}
\acrodef{JFLAP}[JFLAP]{Java - Formal Languages and Automata Package}
\acrodef{LEDR}[LEDR]{Level-Encoded Dual-Rail}
\acrodef{LFP}[LFP]{Local Field Potential}
\acrodef{LIF}[LIF]{Leaky-Integrate and Fire}
\acrodef{LLC}[LLC]{Low Leakage Cell}
\acrodef{LNA}[LNA]{Low-Noise Amplifier}
\acrodef{LPF}[LPF]{Low-Pass Filter}
\acrodef{LRS}[LRS]{Low-Resistive State}
\acrodef{LSM}[LSM]{Liquid State Machine}
\acrodef{LTD}[LTD]{Long Term Depression}
\acrodef{LTI}[LTI]{Linear Time-Invariant}
\acrodef{LTP}[LTP]{Long Term Potentiation}
\acrodef{LTU}[LTU]{Linear Threshold Unit}
\acrodef{LUT}[LUT]{Look-Up Table}
\acrodef{LVDS}[LVDS]{Low Voltage Differential Signaling}
\acrodef{MCMC}[MCMC]{Markov-Chain Monte Carlo}
\acrodef{MEMS}[MEMS]{Micro Electro Mechanical System}
\acrodef{MIM}[MIM]{Metal Insulator Metal}
\acrodef{MLP}[MLP]{Multi Layer Perceptron}
\acrodef{MOSCAP}[MOSCAP]{Metal Oxide Semiconductor Capacitor}
\acrodef{MOSFET}[MOSFET]{Metal Oxide Semiconductor Field-Effect Transistor}
\acrodef{MOS}[MOS]{Metal Oxide Semiconductor}
\acrodef{MRI}[MRI]{Magnetic Resonance Imaging}
\acrodef{NDFSM}[NDFSM]{Non-deterministic Finite State Machine} 
\acrodef{ND}[ND]{Noise-Driven}
\acrodef{NEF}[NEF]{Neural Engineering Framework}
\acrodef{NHML}[NHML]{Neuromorphic Hardware Mark-up Language}
\acrodef{NIL}[NIL]{Nano-Imprint Lithography}
\acrodef{NMDA}[NMDA]{N-Methyl-D-Aspartate}
\acrodef{NME}[NE]{Neuromorphic Engineering}
\acrodef{NRZ}[NRZ]{Non-Return-to-Zero}
\acrodef{NSM}[NSM]{Neural State Machine}
\acrodef{OTA}[OTA]{Operational Transconductance Amplifier}
\acrodef{PCB}[PCB]{Printed Circuit Board}
\acrodef{PCHB}[PCHB]{Pre-Charge Half-Buffer}
\acrodef{PE}[PE]{Phase Encoding}
\acrodef{PCM}[PCM]{Phase Change Memory}
\acrodef{PFM}[PFM]{Pulse Frequency Modulation}
\acrodef{PR}[PR]{Production Rule}
\acrodef{PSC}[PSC]{Post-Synaptic Current}
\acrodef{PSTH}[PSTH]{Peri-Stimulus Time Histogram}
\acrodef{QDI}[QDI]{Quasi Delay Insensitive}
\acrodef{RAM}[RAM]{Random Access Memory}
\acrodef{RELU}[ReLu]{Rectified Linear Unit}
\acrodef{RLS}[RLS]{Recursive Least-Squares}
\acrodef{RMSE}[RMSE]{Root Mean Squared-Error}
\acrodef{RMS}[RMS]{Root Mean Squared}
\acrodef{RNN}[RNN]{Recurrent Neural Network}
\acrodef{ROLLS}[ROLLS]{Reconfigurable On-Line Learning Spiking}
\acrodef{RRAM}[R-RAM]{Resistive Random Access Memory}
\acrodef{SAC}[SAC]{Selective Attention Chip}
\acrodef{SCX}[SCX]{Silicon CorteX}
\acrodef{SD}[SD]{Signal-Driven}
\acrodef{SEM}[SEM]{Spike-based Expectation Maximization}
\acrodef{SLAM}[SLAM]{Simultaneous Localization and Mapping}
\acrodef{SOC}[SOC]{System-On-Chip}
\acrodef{SOTA}[SOTA]{State-Of-The-Art}
\acrodef{SOI}[SOI]{Silicon on Insulator}
\acrodef{SRAM}[SRAM]{Static Random Access Memory}
\acrodef{STDP}[STDP]{Spike-Timing Dependent Plasticity}
\acrodef{STD}[STD]{Short-Term Depression}
\acrodef{STP}[STP]{Short-Term Plasticity}
\acrodef{STT-MRAM}[STT-MRAM]{Spin-Transfer Torque Magnetic Random Access Memory}
\acrodef{STT}[STT]{Spin-Transfer Torque}
\acrodef{SNN}[SNN]{Spiking Neural Network}
\acrodef{SW}[SW]{Software}
\acrodef{TCAM}[TCAM]{Ternary Content-Addressable Memory}
\acrodef{TFT}[TFT]{Thin Film Transistor}
\acrodef{USB}[USB]{Universal Serial Bus}
\acrodef{VHDL}[VHDL]{VHSIC Hardware Description Language}
\acrodef{VLSI}[VLSI]{Very Large Scale Integration}
\acrodef{VOR}[VOR]{Vestibulo-Ocular Reflex}
\acrodef{WTA}[WTA]{Winner-Take-All}
\acrodef{WCST}[WCST]{Wisconsin Card Sorting Test}
\acrodef{XML}[XML]{eXtensible Mark-up Language}
\acrodef{divmod3}[DIVMOD3]{divisibility of a number by three}
\acrodef{hWTA}[hWTA]{Hard Winner-Take-All}
\acrodef{sWTA}[sWTA]{soft Winner-Take-All}

\acrodef{ORN}[ORN]{olfactory receptor neuron}
\acrodef{OR}[OR]{olfactory receptor}
\acrodef{SNR}[SNR]{signal-to-noise ratio}

\newcommand{\NORCO}{N_\text{ORCO}}
\newcommand{\NOR}{N_\text{OR}}
\newcommand{\popen}{p_\text{open}}
\newcommand{\pclose}{p_\text{close}}
\newcommand{\Tdata}{T_\text{data}}
\newcommand{\Ttotal}{T_\text{total}}
\newcommand{\dt}{\text{dt}}
\newcommand{\Rgap}{R_\text{gap}}
\newcommand{\Cdl}{C_\text{dl}}
\newcommand{\Rt}{R_\text{t}}

\begin{document}

\title[K.~Max et al.]{Synthetic Biology Meets Neuromorphic Computing: Towards a Bio-Inspired Olfactory Perception System}

\author{Kevin Max$^{1,2,\dagger}$, Larissa Sames$^{3}$, Shimeng Ye$^{4}$, \\  Jan Steink\"uhler$^{3,5,\dagger}$, and Federico Corradi$^{4,\dagger}$}

\address{$^{1}$Neural Computation Unit, Okinawa Institute of Science and Technology, Japan}
\address{$^{2}$Department of Physiology, Bern University, Switzerland}
\address{$^{3}$Bio-Inspired Computation, Institute of Electrical and Information Engineering, Kiel University, Kiel, Germany}
\address{$^{4}$Electrical Engineering, Eindhoven University of Technology, The Netherlands}
\address{$^{5}$Kiel Nano, Surface and Interface Science KiNSIS, Kiel University, Kiel, Germany}
\address{$^{\dagger}$Corresponding authors}
\ead{kevin.max@oist.jp, jst@tf.uni-kiel.de, f.corradi@tue.nl}
\vspace{10pt}
\noindent{\it Keywords}: neuromorphic engineering, olfaction, chemical sensing, synthetic biology

\begin{abstract}
In this study, we explore how the combination of synthetic biology, neuroscience modeling, and neuromorphic electronic systems offers a new approach to creating an artificial system that mimics the natural sense of smell. We argue that a co-design approach offers significant advantages in replicating the complex dynamics of odor sensing and processing.
We propose a hybrid system of synthetic sensory neurons that provides three key features: a) receptor-gated ion channels, b) interface between synthetic biology and semiconductors and c) event-based encoding and computing based on spiking networks.
Our approach is validated using simulation-based modelling of the complete sensing and processing pipeline.
This research seeks to develop a platform for ultra-sensitive, specific, and energy-efficient odor detection, with potential implications for environmental monitoring, medical diagnostics, and security.
\end{abstract}

%
%
%
%
%

\section{Introduction}
\label{sec:intro}


Biological olfaction is a remarkably efficient sensory system, surpassing engineered chemical detection methods through redundant neural circuits that ensure robust responses~\cite{yang2022redundant}, combinatorial coding that enables the detection of multitude of odors
~\cite{Bushdid2014,Gerkin2015,fonollosa2012quality}, and exceptional sensitivity~\cite{laska2014olfactory}. These advantages have inspired extensive efforts to replicate biological olfaction, leading to significant improvements in artificial olfaction ~\cite{raman2011mimicking,brattoli2011odour}. Today’s electronic noses (e-noses) and artificial olfactory biosensors employ arrays of chemosensors and advanced pattern recognition algorithms, enabling odor classification in an impressive variety of fields, from biomedical signal analysis~\cite{wasilewski2022olfactory} and environmental monitoring~\cite{prasad2022electronic}, to food safety~\cite{lu2022electronic}
, agricultural productivity~\cite{kiki2024bibliometric}, and security~\cite{pobkrut2016sensor}.

Despite these notable achievements, e-nose technologies continue to face several key challenges that have slowed their broader adoption. These include: (i)~sensor drift, which deteriorates performance over time~\cite{bosch2022electronic}; (ii)~cross-sensitivity, which leads to reduced specificity when sensors respond to multiple volatile compounds~\cite{rabehi2024advancements}, in contrast to the modulation of the combinatorial code during the perception of a mixture of odorant observed in biology~\cite{de2020modulation}; (iii)~in the case of widely used metal oxide (MOx) sensing elements: significant power consumption due to high operating temperatures~\cite{Schmuker2022};
(iii)~limited computational capacity for interpreting the complex combinatorial patterns inherent in real-world odors, as many e-nose systems rely on embedded microprocessors as the core control
unit with limited processing capacity~\cite{cheng2021development}; and (iv)~insufficient adaptability, as current systems cannot dynamically refine their responses from experience, unlike their biological counterparts~\cite{raman2011mimicking}. Observing how biological olfaction circumvents these problems suggests a promising model for the next generation of artificial olfactory devices, particularly given its robust signal processing capabilities, cross-selective receptor integration, and the ability to learn from repeated exposure~\cite{szulczynski2018determination, lee2023olfactory, imam2020rapid}.

A striking feature of biological olfaction is its capacity to handle the vast complexity of real-world odors. In nature, few odorants appear in isolation and unpredictable interactions among volatile compounds often challenge artificial sensors. In contrast, humans, for example, can distinguish an astonishing large number of odors; an early estimate puts the number around 10,000, although subsequent studies point to a significantly higher figure (albeit with some debate)~\cite{Bushdid2014,Gerkin2015}.

An important question is how olfactory receptor responses are distributed across different odorants.
Until recently, consensus has been that individual receptors always respond to a range of volatile compounds~\cite{Hallem_2006,Pelz_2006,ma2012distributed,munch_door_2016}, implying that odors compete for overlapped receptor sites through a process often described as odor occlusion~\cite{cleland2012sequential,xu2023odor}.
Such competitive interactions can be highly non-linear: one odorant can amplify receptor activity, while another can dampen it, leading to outcomes that defy simple additive models~\cite{xu2023odor}.
However, as has recently been argued by Wachowiak et al.~\cite{wachowiak2025recalibrating}, these findings are mostly based on stimulation protocols which exceed naturally relevant concentration ranges by orders of magnitude.
In fact, under naturally occurring concentrations, tuning curves of receptor neurons appear highly odorant-specific~\cite{zhang2013design,conway2024perceptual,dennler2025neuromorphic}, in some cases activating only a single glomeruli~\cite{burton2022mapping}.
Furthermore, a large meta analysis study~\cite{wachowiak2025recalibrating} suggests that naturally occurring odor concentrations are often found near or below their detection threshold. 
Such view introduce many layers of complexity, making biological olfaction remarkably adept at navigating real-word odor spaces. Unlocking and understating those principle in engineering systems might push artificial olfaction beyond its current limits and toward robust, adaptable and context-aware sensing technologies.

\subsection{Background \& state-of-the-art}

\subsubsection{Sensing technology.} Bioinspired artificial olfactory systems use various sensing technologies, including metal oxide semiconductors, conductive polymers, and integrated MEMS sensor arrays
~\cite{s120709635,raman2008bioinspired,Schmuker2022}. These systems operate by detecting changes in physical properties, primarily electrical conductance, when volatile compounds interact with sensing materials. Discriminatory power is achieved through differential molecular interactions between multiple sensors and pattern recognition of the resulting combinatorial responses~\cite{koickal2007analog,covington2007towards,bernabei2012large,tang2010development,kim2024ultralow}.

Metal oxide sensors, while effective for chemical detection, are limited both by their rather slow response time (on the order of many seconds), and drift over longer timescales~\cite{Dennler2024,wilson1997signal}. Drift manifests as temporal shifts in sensor response under identical working conditions, primarily caused by chemical and physical interactions at the sensing film microstructure, as well as external factors such as temperature and humidity variations.
For example, recent analysis of the widely used gas sensor dataset by Vergara et al.~\cite{vergara_chemical_2012} revealed how sensor drift can significantly affect recordings, hindering realistic assessment of sensor performance~\cite{Dennler2024}.
Although traditional approaches attempt to address drift through regression models and environmental compensation~\cite{pan2023comprehensive,abdullah2022correction}, bio-inspired neuromorphic implementations offer promising alternatives through spike-based encoding~\cite{vanarse2017investigation} and adaptive learning mechanisms that could potentially provide stability and resistance to drift. For instance, Linster and Cleland in~\cite{borthakur2019spike} model sensor inputs that are first normalized via inhibition in the glomerular layer, ensuring a concentration-invariant encoding. Synaptic plasticity based on spike-timing-dependent plasticity in the external plexiform layer then enables the system to continually adapt its internal representations of odors. This allows the network to refine its response profiles over time, compensating for both baseline shifts and changes in sensor responsiveness through experience-driven learning. Such neuromorphic approaches aim to reduce the complexity of the data and implement adaptive algorithms that can follow changing conditions, potentially offering a more robust solution to the persistent challenge of sensor drift.


Building on these concepts, integrated analog on-chip learning circuitry has demonstrated efficacy in odor detection and classification. For example, Covington et al.~in~\cite{covington2007towards} used an 80 element chemoresistive microsensor array combined with a microfluidic package, while Tang et al.~in~\cite{tang2010development} relied on commercially available sensors and a traditional signal processing pipeline in a microprocessor to distinguish three fruit fragrances. In a related work~\cite{bernabei2012large}, Bernabei et al.~utilized a large array of chemosensors with slightly varied sensor elements to emulate the overlap selectivity characteristic of biological olfaction, generating rich combinatorial responses. More recently, Rastogi et al.~\cite{rastogi2024neuromorphic} introduced a neuromorphic analog e-nose that directly encodes the gas concentration using MOx sensors and an analog front-end that transforms sensor responses into spike-based representations. Their approach leverages the time difference between spikes in two distinct processing pathways to infer gas concentration, reducing data redundancy, and improving power efficiency. 

However, current polymer and metal oxide-based sensing elements remain relatively simple compared to biological olfactory systems, facing challenges with odor specificity and stability~\cite{dung_applications_2018,virumbrales2024sensory}.
Although these artificial sensors can detect certain volatile compounds, they often require high operating temperatures and exhibit limited effectiveness in discriminating complex molecular mixtures~\cite{Schmuker2022,el2021overview}.
A new generation of nanomaterials-based MOx sensors, and large-scale monolithically integrated nanotube sensor arrays~\cite{wang2024biomimetic}, promises to alleviate some of these issues, but current implementations are still in their prototype phase~\cite{Abideen2024May,rabehi2024advancements}. 
Moreover, their continuous sampling needs and high power consumption stand in stark contrast to the low-power, event-driven nature of biological olfaction~\cite{raman2011mimicking}. These fundamental discrepancies in functionality and energy efficiency highlight the limitations of conventional chemosensors in achieving the richness and adaptability seen in biological odor detection.

Compared to traditional chemosensors, biosensors directly interface with a biological element, such as a whole cell or a purified protein, which is then linked to a physical transducer~\cite{glatz2011mimicking}. Although harnessing authentic biological receptors offers the potential for greater specificity, it also brings practical hurdles, including potentially shorter operating lifetimes, contamination risks, and more complex maintenance requirements. A promising alternative is to adopt a bottom-up synthetic biology strategy that extracts specific components of the olfactory system, like receptor proteins, and deploys them in controlled, nonliving environments. For example, reconstitution of mammalian and insect olfactory receptor proteins into stable platforms has been achieved in combination with electrical, optical, and acoustic signal transduction methods~\cite{lu2022insect}. However, early efforts often focused on single receptors in isolation, which failed to leverage the amplification cascades inherent to living cells.

Notably, some more recent works have demonstrated a more complete reconstitution of the olfactory system, in particular, into bio-mimetic environments, for example membrane bound vesicles formed by membrane extraction, or virus-like particles~\cite{hirata2021biohybrid,cheema2021insect,khadka2019ultrasensitive,yamada2021highly}.
These approaches leverage the fact that insect olfaction is largely based on multivalent interactions between the conserved co-receptor Orco and the odorant binding subunits (odorant receptors, \acs{OR}s), which together form a ligand-gated ion channel. Khadka et al.~\cite{khadka2019ultrasensitive} have demonstrated successful insertion of multiple odorant receptors into surface-anchored lipid vesicles.
Yamada et al.~\cite{yamada2021highly}~have demonstrated parts per billion sensitivity of an Orco-OR complex inserted into a lipid bilayer to octenol, a biomarker in human breath. Detection of femtomolar concentrations of VUAA1, E2-hexenal, 4-ethylguaiacol molecules was demonstrated using odorant receptors Or10a, Or22a, and Or71a~\cite{khadka2019ultrasensitive}.

These studies show the feasibility of reconstituting functional and highly sensitive insect ORs in a non-living minimal system with electrical readout. However, none of these works have considered the multiplexed decoding signal from multiple ORs, the hallmark of the olfactory system. Additionally, previously used  transducers, e.g.~impedance spectroscopy, require energy-intensive sampling, and thus do not fully harness the event-driven character and amplification capabilities of biological olfaction. 

\subsubsection{Decoding algorithms for odor discrimination.}
The signal restoration challenge is being recognized as one of the central problems in neuromorphic olfaction~\cite{marco2012signal,persaud2013engineering, raman2011mimicking}.
In fact, for more than 20 years, researchers have concluded that the speed and accuracy of odor recognition in insects cannot be attributed solely to the performance of olfactory receptors due to their slow time constant and high variability. For instance, insects obtain fast, accurate odor recognition by transforming the slow, variable receptor signals in the antennal lobe by means of its network dynamics that converts these inputs into richer spatiotemporal patterns, especially during the initial transient, that carry far clearer, odor-specific information~\cite{muezzinoglu2009chemosensor}.

The requirement to correctly classify odor concentrations poses additional demand on the analyzing algorithms, as do signal occlusion and noise effects. 
Thus, several models of signal processing downstream of the sensor have been investigated. Examples of such models includes the architecture of the mammalian olfactory bulb~\cite{imam2020rapid}, insect models that focus at the receptor level~\cite{nowotny2014drosophila,schmuker_predicting_2007}, within the antennal lobe~\cite{muezzinoglu2009chemosensor} or the macroglomerual complex~\cite{pearce2013rapid}; and neuromorphic network that replicate insect coding~\cite{jurgensen2021neuromorphic,schmuker_2014, BetkiewiczENEURO.0305-18.2020} and learning~\cite{zhao_predictive_2021} strategies. A comparative study even contrasts these biologically grounded approaches with classical signal processing strategies~\cite{ratton_comparative_1997}, highlighting the advantages of spatio-temporal codes for robust odor classification.

Given this rich literature we can roughly group methods for odor signal analysis into two classes: data analysis tools (e.g.,~principal component analysis, support vector machines) and biorealistic network models.
While data analysis tools have been used for odor classification since many years~\cite{ratton_comparative_1997,nowotny2014drosophila},
biologically realistic network models are now studied with the goal of reaching the accuracy and efficiency of the animal olfactory system~\cite{pearce2013rapid,BetkiewiczENEURO.0305-18.2020,imam2020rapid,jurgensen2021neuromorphic}.
Many of these models perform classification using a spatiotemporal code while learning in an unsupervised fashion.
Additionally, some studies attribute the efficiency of natural systems to two main intrinsic mechanisms of biorealistic neural networks (i.e., \ac{SNN}): \textit{i)} inhibition (which triggers a winner-takes-all competition~\cite{schmuker_2014,pearce2013rapid,imam2020rapid}) and \textit{ii)} temporal integration (due to neuron and network dynamics)~\cite{muezzinoglu2009chemosensor, imam2020rapid}.
Recent work has also pursued the implementation of learning from teaching signals (supervised and reinforcement learning) in models of the insect olfactory system, in particular for associative learning~\cite{zhao_predictive_2021,10.1162/neco_a_01615}.

\subsubsection{Neuromorphic implementation of olfactory models.}
Several hardware implementations have been demonstrated in digital Field Programmable Gate Arrays (FPGA) hardware~\cite{pearce2013rapid},  digital neuromorphic devices~\cite{imam2020rapid,schoepe2024odour} and analog neuromorphic chips~\cite{koickal2007analog,schmuker_2014,10.3389/fnins.2013.00011,rastogi2024neuromorphic}.
Recently, a bio-realistic network~\cite{imam2020rapid}, based on the architecture of the mammalian olfactory bulb, was implemented on the Loihi neuromorphic digital chip, demonstrating the benefits of employing an algorithm based on online learning and spike timing.
However, the system was trained using a separately recorded data set of pure chemicals~\cite{vergara_chemical_2012}, therefore lacking integration with real chemosensors, and criticism has been raised regarding the generalizability of the network model~\cite{Dennler2024}. 
Beyond odor classification, recent developments have also tackled the challenge of odor localization; Schoepe et al.~\cite{schoepe2024odour} demonstrated a fully event-driven neuromorphic implementation using stereo arrays of MOx sensors interfaced with a spiking neural network on SpiNNaker hardware~\cite{painkras2013spinnaker}. Their approach effectively leveraged a spike-time-difference encoding strategy to infer odor source direction, while Dennler et al.~\cite{dennler2025neuromorphic} emphasized the effectiveness of neuromorphic event-driven sensing principles for efficiently capturing sparse and dynamic odor signals encountered in turbulent real-world environments.

The neuromorphic olfactory sensor literature indicates that there is considerable scope for improving neuromorphic chemosensory systems. There are several limitations in many of the current implementations, such as the lack of signal conditioning, the lack of integration of sensing and decoding in a single platform, high power consumption, and the poor sensitivity of the artificial sensor array.
So far, the only truly integrated neuromorphic sensing platform for olfaction has been developed by Koickal et al.~in~2006~\cite{koickal_analog_2006,koickal2007analog}.
To our knowledge, this platform did not leave the proof-of-concept stage, and subsequent work by the authors has used conventional processing hardware~\cite{pearce2013rapid}.
Our present work is focused on integrating hardware realizations of all stages into one single system.

\section{Co-design of synthetic biology, bio-inspired models, and neuromorphic systems} 

\begin{figure}
    \centering
    \begin{subfigure}[b]{0.49\textwidth}
        \centering
        \raisebox{-0.5\height}{\includegraphics[width=10cm]{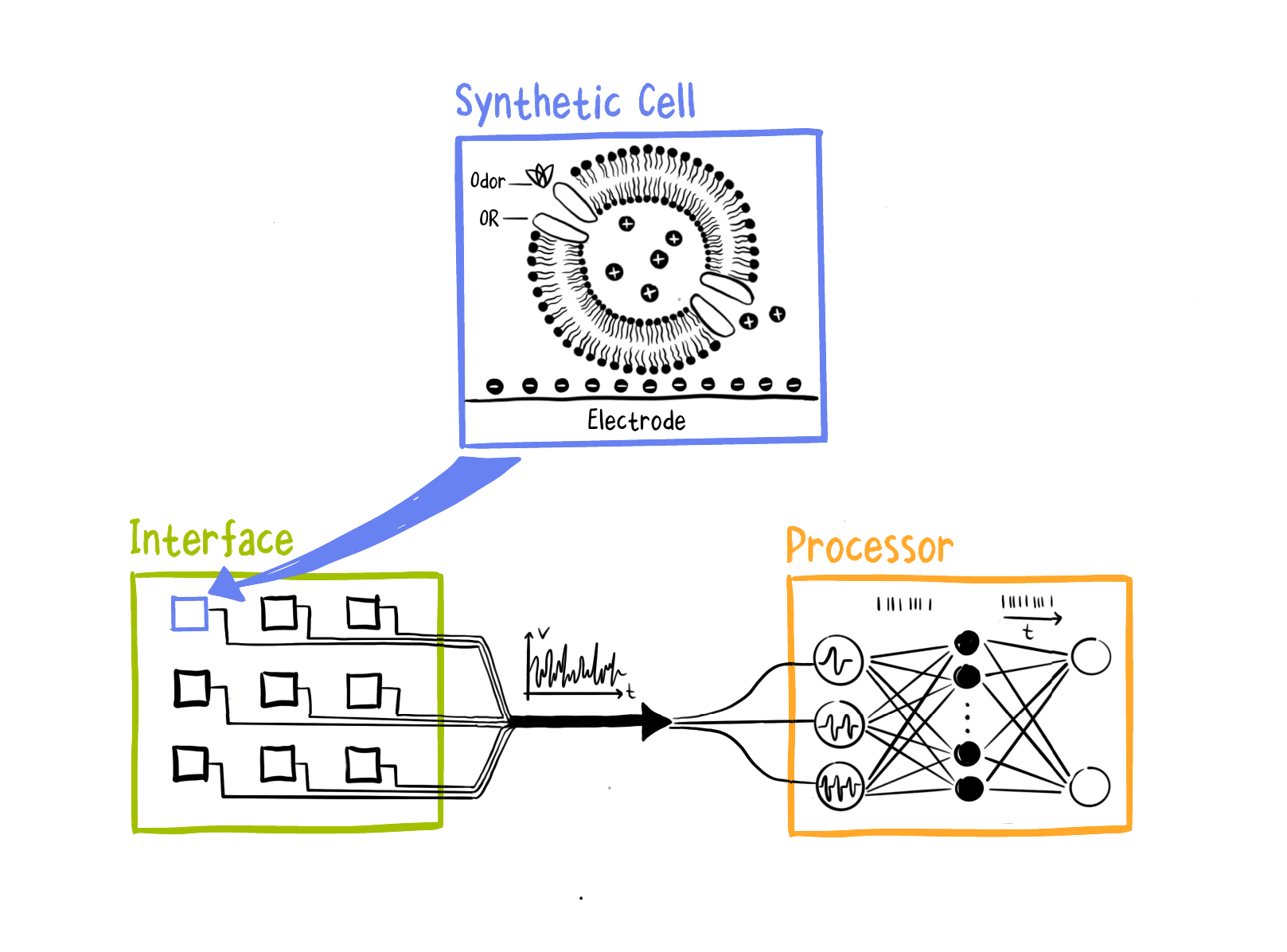}}
    \end{subfigure}
    \hfill
    \begin{subfigure}[b]{0.49\textwidth}
        \centering
        \raisebox{-0.5\height}{\includegraphics[width=\textwidth]{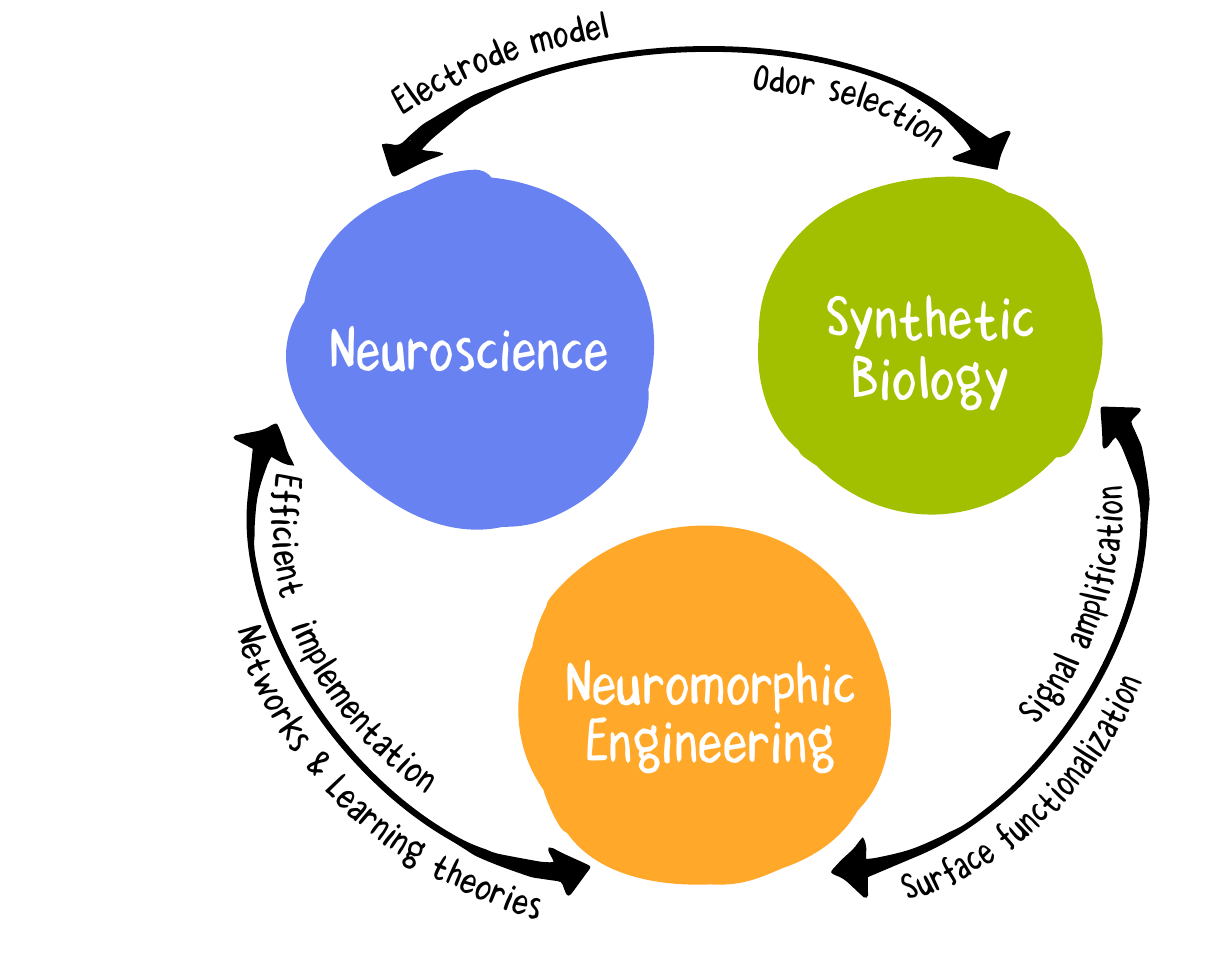}}
    \end{subfigure}
    \caption{\textit{Left:} Bio-inspired information flow from odor binding, depolarization of a synthetic cell, spike generation in a CMOS circuit (hybrid synthetic sensory neuron) and processing in a spiking neural network.  \textit{Right:} Co-design of SYNCH: ``Combining SYnthetic Biology \& Neuromorphic Computing for CHemosensory perception'', interconnecting the three disciplines of sensing with synthetic biology, efficient bio-inspired learning models, and neuromorphic hardware engineering.}
    \label{fig:co-design}
\end{figure}

A comprehensive approach is crucial in developing advanced neuromorphic chemosensory systems that effectively integrate sensor arrays, decoding algorithms, and their physical realization~\cite{hurot_bio-inspired_2020}. In this work, we present a novel methodology and preliminary results on how to move beyond the \ac{SOTA} by addressing the lack of codesign between these three components.
Unlike previous chemical detection systems inspired by the olfactory pathway, which have treated sensing \cite{covington2003combined,nowotny2014drosophila}, signal processing \cite{lozowski2004signal,caticha2002computational}, and neuromorphic computation \cite{imam2020rapid} as separate components, our approach integrates these aspects to tackle the challenge of chemosensory perception in a comprehensive way. 
In this work, we introduce a new concept to couple the depolarization of synthetic cells made from biological materials directly with CMOS electronic circuits designed for spike generation, we call these new class of bio-electronic devices hybrid synthetic sensory neurons, and we depict them in Fig. \ref{fig:co-design}.


In doing so, our framework harnesses the advantages of co-design in a joint environment. Here, the three disciplines (neuroscience, synthetic biology,  neuromorphic engineering) are developed together and connected to each other (see Fig.~\ref{fig:co-design}):

\begin{itemize}
\item The bottom-up sensor design using synthetic biology informs the \ac{OR} and electrode response model, allowing us to simulate electrode responses with a biologically realistic ion channel model.
In turn, this sets realistic estimates for electrode responses, which are crucial for the design of sensor and amplification stages.
Importantly, this also allows us to pick Orco-OR complexes which respond strongly to the set of odor combinations that we aim to detect, while minimizing the number of required \acs{OR}s which need to be deployed.
\item The design of the sensor and neuromorphic processing stages are naturally intertwined: the bio-electro interface needs to be carefully selected and tuned (e.g.~with regard to the electrode surface material, synthetic cell immobilization technique, geometry of sensor arrays, odorant flushing, and amplification electronics) to allow for optimal readout of the ionic currents, aiming for fast odorant throughput, high \ac{SNR} and long lifetime of the immobilized synthetic cells. 
\item Computational theory and neuromorphic hardware in turn inform each other about how to realize energy-efficient olfactory processing and learning under real-life conditions.
Theoretical ideas of few-shot, online learning in noisy environments can be tested on spiking hardware, and need to be made robust against constraints such as sparse coding and weight quantization.
Here, a central issue is the compensation of sensor drift, which needs to be addressed by a theoretical approach, e.g.~by a homeostatic learning mechanism inspired by biology, and then tested using the full sensor/processing hardware.
\end{itemize}

Our approach is not without challenges, as individual parts need to be optimized and tuned together but must still allow for the necessary flexibility. This flexibility is needed as knowledge and research about the individual parts progresses simultaneously. The parallel approach is, on one hand, due to constraints of scientific funding—collaborative grants typically work on a common timeline from proposal to project end. However, parallelized development is also, in general, desired for time efficiency. As such, the collaborative tools we develop, particularly those that bridge data-driven synthetic biology with electrical and information engineering, not only represent standalone contributions to each field, but might also be useful in other applications.
In the following, we detail our co-design approach, validate our sensing pipeline using simulation-based modelling, and discuss ongoing challenges.



\section{Results}



  \subsection{Insect odor receptors provide a modular sensing platform}
Insect olfaction is largely based on transmembrane protein complexes consisting of an odorant receptor (OR) and a co-receptor, Orco. While ORs diverge in their sequence and odor specificity between the multitude of insects and the ecological niches they inhabit, the co-receptor Orco is highly conserved across insect species. Several experiments show that it is possible to create functional receptor complexes by combining Orco of one species with a foreign OR \cite{zhao2024structural,jones2005functional}. From an engineering point of view, this suggests the possibility of a modular platform in which the base assembly of an Orco-based sensing element allows the swapping of ORs with different response profiles to target various compounds. To demonstrate this modularity, we set out to computationally assemble Orco-OR complexes using AlphaFold modeling~\cite{yu2023alphapulldown,jumper2021highly}.

\begin{figure}[p]
    \centering
    \includegraphics[width=0.8\textwidth]{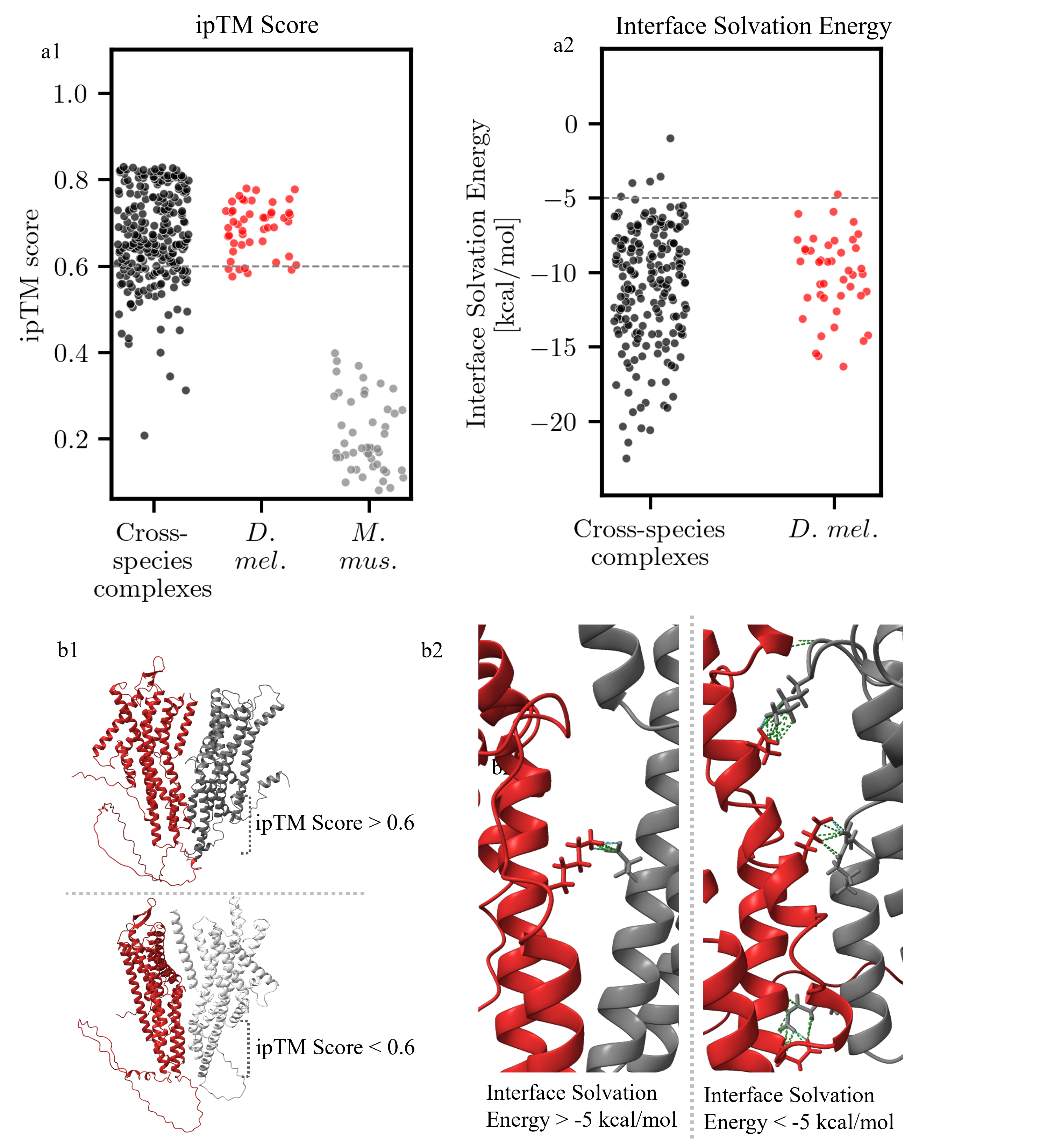}
    \vspace{2em}
        \caption{AlphaFold predictions suggest hundreds of possible stable complexes between \textit{D.~melanogaster} Orco and foreign odor receptors (OR). \textit{a1:}~AlphaFold ipTM score for Orco-OR complexes formed from \textit{D.~melanogaster} Orco combined with OR of different insect species (black), and combined with \textit{D.~melanogaster} (Dm) ~OR (positive control, red) and \textit{M.~musculus}~OR~(negative control, grey). \textit{a2:}~Interface solvation energy prediction for Dm with foreign OR above the ipTM cut-off of 0.6. Dashed grey lines indicates cut-off thresholds.
    \textit{b1:}~Dm Orco (red) dimerization with \textit{M.~sexta} OR4 resulted in an ipTM score above 0.6 (dark grey) while the dimerization with \textit{L.~migratoria} OR113 led to a score below 0.6 (grey). \textit{b2:}~ Dm Orco and \textit{B.~mori} OR15 complex with favorable solvation energy ($> -5$ kcal/mol), whereas the dimerization with \textit{C.~quinquefasciatus} OR151 resulted in a low ($< -5$ kcal/mol) solvation energy. Varying number of bonds can be observed in the anchor domain (brackets). Green dashed lines indicate electrostatic interactions, while dashed blue lines represents hydrogen bonds.
    }
    \label{fig: ipTM_and_interface_solvation_energy}
\end{figure}
Specifically, we selected the Orco variant from the \textit{D.~melanogaster} (Dm) fruit fly and considered a total of 348 Orco-OR complexes, including non-Dm ORs. Foreign ORs derived from either other insect species or from a mammal (mouse, \textit{M.~musculus}) as a negative control. Each predicted Orco-OR structure was evaluated using the interface-predicted template modeling (ipTM) score, with higher values indicating greater confidence in the prediction accuracy of the protein-protein interface. An empirical cut-off point of 0.6 was chosen (Figure \ref{fig: ipTM_and_interface_solvation_energy}a1). The protein-protein interface refers to the region where the two proteins directly interact with each other. The interface of the Alphafold-predicted Orco-OR complexes is of great interest as it provides insight into the plausibility and structural integrity of the possible receptor complex.~\cite{evans2021protein, jumper2021highly}. As expected, the positive control, consisting only of Orco-OR complexes from the same species (Dm), yielded mostly acceptable modeling outputs, as reflected in ipTM scores above 0.6. In contrast, the negative control, which used the mammalian OR that are not expected to form  complexes with Orco, did not produce any acceptable models (Figure \ref{fig: ipTM_and_interface_solvation_energy}a1). Interestingly, for Dm Orco and ORs from foreign species, we found a large fraction of potentially accurate models with ipTM scores above the 0.6 cut-off. This suggests that AlphaFold models are relevant for understanding the structures of Dm Orco in complex with ORs from foreign species.

To further evaluate the structures, we considered a second metric: interface solvation energy, which provides insight into the stability of the non-covalent binding between Orco and OR (Figure \ref{fig: ipTM_and_interface_solvation_energy}a2). More negative values indicate a more favorable binding interface~\cite{kraml2019solvation}. We found that the obtained values for foreign OR were similar to those of the native Dm Orco-OR complexes (Fig.~\ref{fig: ipTM_and_interface_solvation_energy}b2), indicating comparable complex stability even for foreign ORs. Furthermore, we manually inspected the most favorable predictions for each group. In both cases, the interactions between the two subunits were confined to a segment of helical bundles stabilized by hydrogen bonds and electrostatic interactions located away the membrane interface (Fig.~\ref{fig: ipTM_and_interface_solvation_energy}a2), consistent with the anchor domain and interactions identified in experimental studies~\cite{Butterwick2018,wang2024structural,zhao2024structural}. These results suggest that the OR structures generated by AlphaFold are promising candidates for computational exploration of the wide diversity of ORs.

\begin{figure}[t]
    \centering
    \begin{subfigure}[t]{0.49\textwidth}
        \centering
        \includegraphics[width=\textwidth]{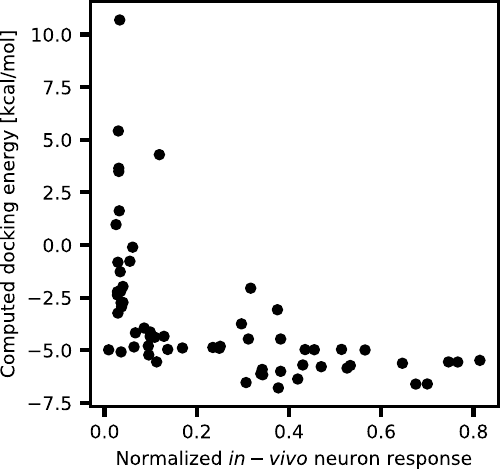}
    \end{subfigure}
    \hfill
    \begin{subfigure}[t]{0.47\textwidth}
        \centering
        \includegraphics[width=\textwidth]{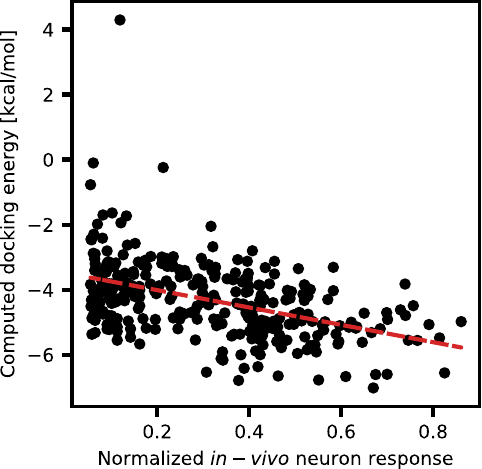}
    \end{subfigure}
    \caption{Comparison of ligand binding energies of Dm - Or69a, Or98a, Or19a and Or92a. Each datapoint corresponds to one odor molecule. Left: Terpene subset. Right: All odors from the DoOR dataset with neuron response above the spontaneous firing rate (SFR). Dashed line shows linear regression with $R^2 = 0.2$.}
    \label{fig:ligand_comparison}
\end{figure}

To demonstrate the potential to build a digital representation of OR-odor interactions, we challenged the generated AlphaFold Orco-OR structures with \textit{in-vivo} recordings of neuronal activity. We used the DoOR 2.0.1 dataset~\cite{Galizia2010,munch_door_2016}, a comprehensive database of olfactory responses from various measurement techniques such as single sensillum recordings, calcium imaging, and heterologous expression~\cite{Hallem_2004,Goldman_2005,Pelz_2006,Yao_2005,Hallem_2006,Turner_2009,Galizia2010,Marshall_2010,Silbering_2011,Dweck_2013,Gabler_2013,Ronderos_2014,Dweck_2015,bruyne_odor_1999,de_bruyne_odor_2001,kreher_molecular_2005,schmuker_predicting_2007,naters_receptors_2007,kreher_translation_2008,de_bruyne_functional_2010,stensmyr_conserved_2012}.
In the DoOR dataset, all recordings have been merged and normalized across studies, providing response data for 78 types of sensory units to 692 odorants. Both excitatory and inhibitory responses are included in the dataset, represented as responses above or below the spontaneous firing rate (SFR). The magnitude of the SFR varies between \acs{OR}s and is typically below 0.1 normalized units. As we aim to reconstitute \acs{OR}, we exclude units from the dataset which can not be clearly linked to a single \acs{OR} type, such as responses from multiple receptor types or ionotropic receptors.
This leaves a total of 52 possible~\acs{OR} types as the starting point of our analysis.
Note that the DoOR dataset contains recordings from stimulations at high concentrations not found in nature.
As discussed in Section~\ref{sec:intro} and Ref.~\cite{wachowiak2025recalibrating}, odor occlusion and separability are strongly affected by odor concentration, which will need to be taken into account in real-world deployment of our sensing platform (see Discussion~\&~Outlook).

For each \acs{OR}-odor combination, the DoOR dataset provides a normalized \textit{in-vivo} neuron response (`consensus dataset with global normalization' in~\cite{munch_door_2016}).
If the predicted OR structures are accurate, we expected that neuronal activity would correlate with the binding of the response-inducing molecule to the OR model. We initially focused on terpenes, a class of carbohydrates with a large natural abundance, because we speculated that these molecules were a feasible target for a computationally relatively simple molecular docking approach. We calculated the binding energy of terpenes, to the generated OR structures using Vina Dock (see Materials and Methods) for a subset of Dm ORs and plotted these against the measured neuronal response from the DoOR dataset (Fig.~\ref{fig:ligand_comparison}, left). We found that the difference in binding energy between responding (> 0) and non-responding neurons ($\leq$ 0) was highly significant (Student's t-test, $p < 3 \cdot 10^{-7}$), suggesting that the predicted structures have realistic binding pockets for terpenes. Motivated by this, we investigated the same Dm OR subset using all molecules present in the DoOR dataset. Here, we obtained a weak correlation between neuron activation and binding energy (Fig.~\ref{fig:ligand_comparison}, right). At this point it is unclear whether the scatter can be explained by the uncertainty of the molecular docking method or point to more complex molecular recognition mechanisms. Currently, we are investigating an even larger number of ORs and binding molecules with refined computational approaches. If the responses of olfactory neurons to odors are indeed explained simply by the binding energy of an odor to the OR, the molecular recognition process of insect olfaction can be condensed into a measure of this binding energy. Taken together, these results show the potential to use computational tools to predict OR to odor response.

\subsection{Surface stabilization of Orco-OR complexes in synthetic cells}\label{subsec:syn_neuron_interface}
To measure the binding of ORs to odor molecules, OR receptor proteins are typically immobilized on surfaces. To avoid stability problems associated with the immobilization of individual receptors~\cite{hurot_bio-inspired_2020}, we based our system on previous approaches that immobilized lipid membrane vesicles on an electrode surface~\cite{STEINKUHLER20161454,yamada2021highly}. Such vesicles not only provide a native lipid environment but are also able to maintain a membrane potential through ion asymmetry. As in biological sensory neurons, the vesicle membrane potential can be depolarized by Orco-OR channel opening upon ligand binding. In recent years, lipid vesicles with an increasing number of such life-like features are also referred to as synthetic cells and are a field of active developments~\cite{10.7554/eLife.73556}. The increasing number of available synthetic cell modules provides a promising basis to detect the odor to OR binding by event-driven ionic currents, as in biological neurons.

To address the challenge of the necessary co-design approach, we developed a quantitative model of the proposed assembly of lipid-membrane-stabilized Orco-OR ion channels on an electrode. In the model, a fixed number of ion channels generate a current that flows into the electrode impedance. The ion channel was modeled as a resistor with two conductance states, and the electrodes were considered as a parallel resistor-capacitor circuit (see Fig.~\ref{fig:hardware_layout}B Pads \& Equivalent Circuit). The cleft between the synthetic cell and the electrode acts as a resistive element and is relatively large due to the nanoscopic dimensions of the cleft (see Materials and Methods). Analysis of the model showed that the vesicle radius \textit{r} is an important design parameter. The total charge in a synthetic cell scales with~$r^3$, while the number of ion channels scales with~$r^2$. Thus, we expect two regimes with different vesicle sizes: one regime dominated by the finite number of charges encapsulated in smaller vesicles, whereas signals from larger vesicles are sensitive to the density of inserted ion channels. In addition, the overall coverage of signal-generating synthetic cells depends on the ratio of electrode size to vesicle size. Vesicles sizes can be readily tuned between 100~nm and 100~$\mu$m, providing an opportunity to test the different regimes predicted by the model~\cite{yu2009microfluidic}.

\FloatBarrier

\subsection{Selection methodology for odor receptors (OR) and synthetic dataset}

We now discuss our procedure for selecting the odorant and \acs{OR} sets, and how we model sensor responses.
Our approach allows us to answer the following co-design questions:
\begin{itemize}
\item For a given set of odors we want to distinguish, which ORs need to be considered?
\item What magnitude of response signal is expected for an individual synthetic cell?
\item How many synthetic cells of the same receptor type are required for sufficient \ac{SNR}, and which quality is required of the electrode and amplification stages?  
\item Which network models are suited to successfully discriminate sensor data?
\end{itemize}
To this end, we make use of a synthetic dataset based on measurements of odorant responses in Drosophila melanogaster, which we construct from the DoOR dataset.

\begin{figure}[t]
    \centering
    \begin{minipage}{1.0\linewidth}
        \centering
        \raisebox{-0.5\height}{\includegraphics[width=0.31\linewidth]{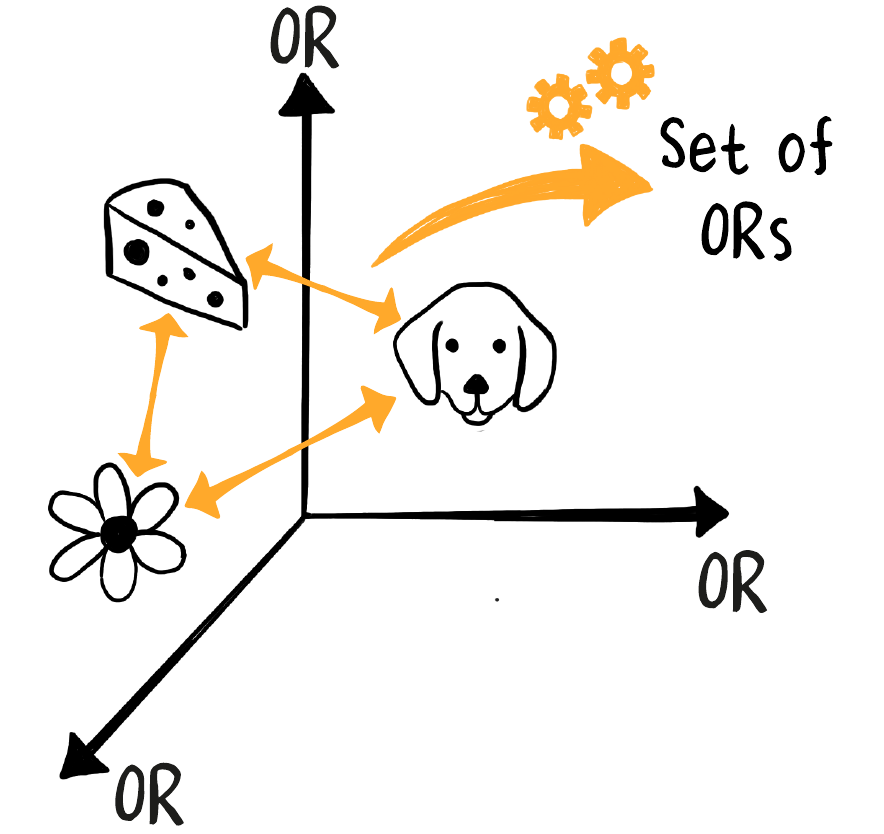}}\hfill
        \raisebox{-0.5\height}{\includegraphics[width=0.46\linewidth]{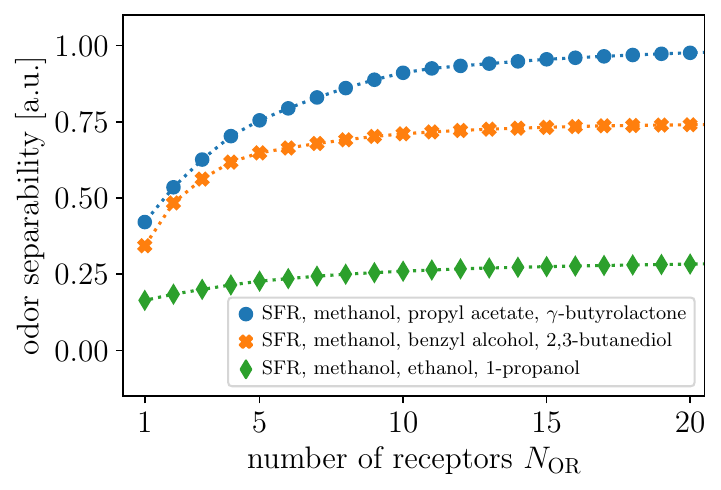}}%
        \raisebox{-0.3\height}{\includegraphics[width=0.21\linewidth]{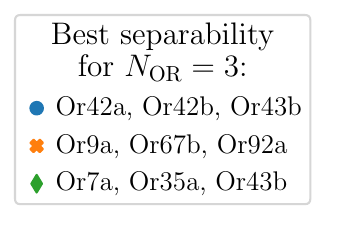}}
    \end{minipage}
    \caption{
    \textit{Left:} Odor separability, exemplified by three smells that occupy a space defined by the responses of different ORs. Our algorithm selects OR for highest separability in the OR/odor space, maximizing the distance between odor responses.
    \textit{Center:} Normalized odor separability vs.~number of Dm \ac{OR}s for different sets of odorants and spontaneous firing rate (no odor, SFR).
    \textit{Right:} \ac{OR}s showing highest separability for each odor set for $\NOR=3$.
    }
    \label{fig:odor_space_analysis}
\end{figure}

We first need to restrict the number of \acs{OR}s and odorants.
The task is to find the minimal number of \acs{OR}s we need to reconstitute in order to reliably differentiate between a given set of odors.
This can be seen as a separation task in the space of responses of all receptors, see Fig.~\ref{fig:odor_space_analysis}~(left).
We achieve this by calculating the `odor separability', i.e.~the Euclidean distance between the response vectors to the odor, and iteratively excluding the \acs{OR} which contributes least to separability;
see Section~\ref{subsec:methods_odor_space_analysis} for a detailed description of the method.

Our analysis yields the odor separability for different sets of odorants vs.~the number of enabled \acs{OR}s, as shown in Fig.~\ref{fig:odor_space_analysis}~(center).
For illustration, we have chosen three sets of odors with differing degrees of relationship in terms of chemical structure (blue, orange, and green sets).
As expected, all odor sets show better separability with larger number of different receptors;
however, relative separability naturally depends on the combination of odors we are probing (note that results are normalized
\textit{across} odorant sets).
Blue: three odorants from different functional groups (respectively: alcohols, esters, and lactones) show large relative separability.
Orange: separability is reduced when considering less diverse functional groups (here: alcohols and aromatic alcohols).
Green: odorants from a single functional group with similar \acs{OR} responses are naturally less separable.
Note that we control against the possibility of no ligand binding at all for a given odorant by including a `no odor' category based on the spontaneous firing rate (SFR); see Section~\ref{subsec:methods_odor_space_analysis} in Materials and Methods.

Most importantly, our algorithm tells us exactly which \acs{OR}s need to be reconstituted for optimal separability of odors and subsequent discrimination of odors by our sensing platform.
As an example, Fig.~\ref{fig:odor_space_analysis}~(right) shows the most relevant \acs{OR}s for $\NOR=3$ for the aforementioned odor sets.

Having selected the odor/\acs{OR} sets of interest, we now construct a synthetic dataset of sensor responses.
To do so, we define a simple ion channel model for individual \acs{OR}s, and subsequently combine their signals in an electrode response model derived from the parameters determined in Section~\ref{subsec:syn_neuron_interface}.

\begin{figure}[t]
    \centering
    \begin{minipage}{1.0\linewidth}
        \centering
        \raisebox{-0.38\height}{\includegraphics[width=0.27\linewidth]{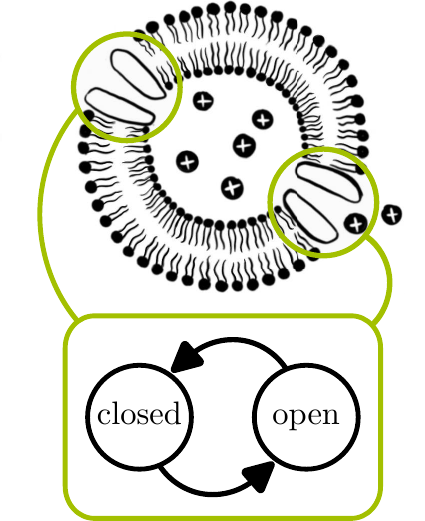}}\hfill
        \raisebox{-0.5\height}{\includegraphics[width=0.73\linewidth]{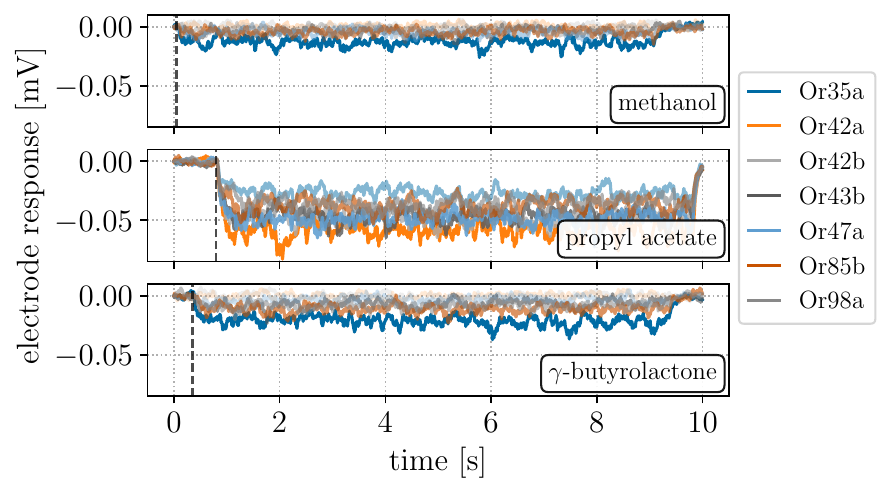}}
    \end{minipage}
    \caption{
    \textit{Left:} Illustration of sensory neuron model. Each Orco-OR complex is modelled as an independent Markov process with two states.
    \textit{Right:}
    Simulated electrode responses to three odorants.
    Each trace shows the response of one synthetic cell comprising 10~Orco-OR complexes of the same receptor type; different receptor types are color-coded.
    We show the 7 most relevant OR types, as determined by our OR reduction algorithm.
    Opacity of traces is scaled with maximum voltage for better readability.
    Odors are only presented during a portion of the overall time frame (dashed vertical line) -- note that odors and onsets are however not correlated in the dataset (multiple presentations of same odor have different onsets).
    }
    \label{fig:synth_dataset_traces}
\end{figure}

We assume that each synthetic cell  contains \acs{OR}s of the same type, and model the response to the binding of an odor molecule as a stochastic process with $\NORCO$ ion channels.
If the ligand binds to a closed Orco channel, the channel opens, and current flows out of the vesicle until stochastic unbinding occurs.
We thus define a Markov model with the opening probability of Orco $\popen$ and closing with $\pclose$, see Fig.~\ref{fig:synth_dataset_traces}~(left).
The open and closed states are converted to currents by comparison with the electrophysiological characterization of VUAA1-induced Orco currents; see Section~\ref{subsec:methods_synth_dataset} and Butterwick et al.~2018~\cite{Butterwick2018}, in particular Extended Data Fig.~2.
Similarly, ion channel noise and low-pass filter properties of the vesicle membrane are modeled on the basis of electrophysiological characterization.
For the electrode response, we assume that synthetic cells of different receptor types are physically separated, such that each electrode probes only one OR type. In addition, we parametrize the electrode model as described in Section~\ref{subsec:syn_neuron_interface}. Analysis of the electrical circuit shows that the sensor response to an odor molecule is given by a low-pass filter of the summed Orco-OR currents in each neuron.
From this procedure, we obtain a time series of voltages with sensor responses per odor--OR type combination, see Fig.~\ref{fig:synth_dataset_traces}.
The model is flexible with respect to which odorants and \acs{OR}s to simulate, supporting the full DoOR~2.0.1 dataset of 692 odors and 52 \acs{OR} types.
In order to increase realism w.r.t. actual measurement conditions, where the timing of odor presentation is not always known precisely, the presentation onset is chosen at random within the full simulation time window. To avoid accidental correlation of onset timing with odors, all simulations use different onsets for each recording, such that onsets are not odorant-specific.

\FloatBarrier
\subsection{Odor discrimination with a spiking neural network model}

\begin{figure}[t]
    \centering
    \includegraphics[width=1.0\linewidth]{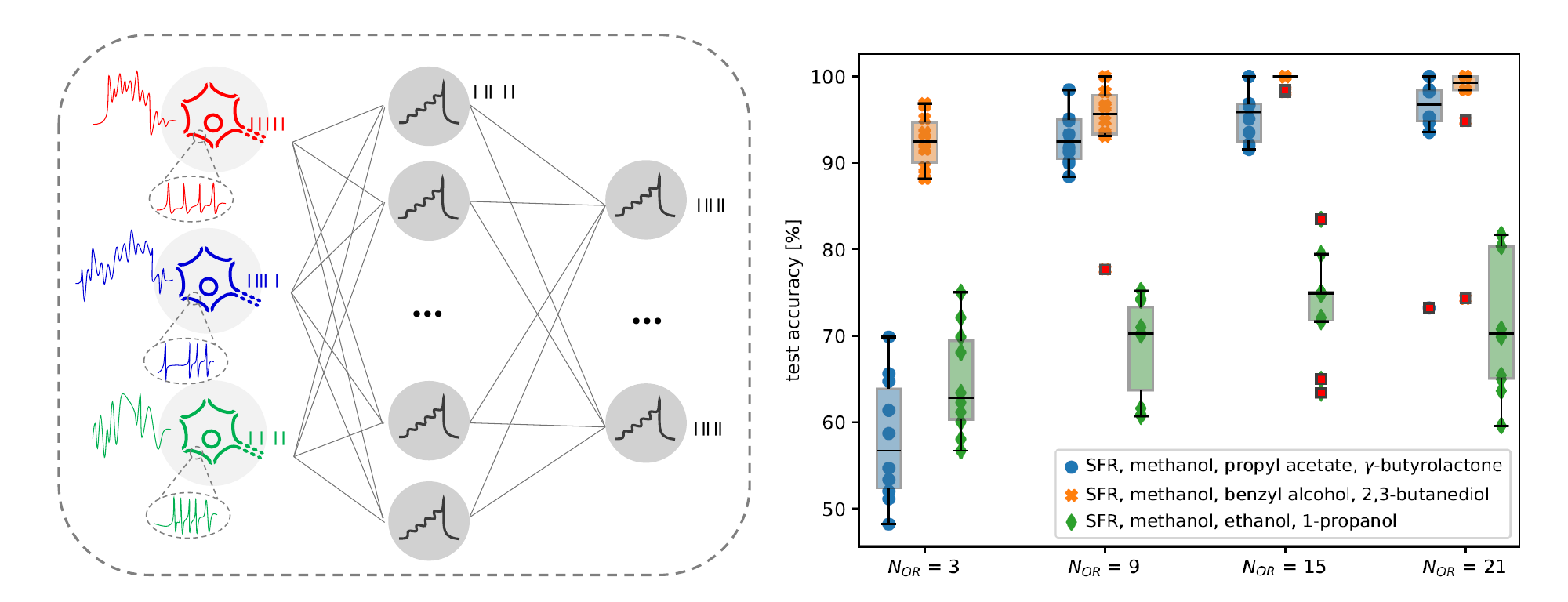}
    \caption{An illustration of the SNN with test accuracy (median and quartiles) for classification of three different sets of odorants (outliers are marked in red squares), with varying numbers of receptors (3, 9, 15, and 21). Each OR group contains 100 samples per odorant, including the ``SFR'' category, resulting in a total of 400 samples. The dataset is split into 70\% training, 15\% validation and 15\% testing. Test accuracy reported corresponds to the epoch with the highest validation accuracy during the 15-epoch training. The number of hidden layer neurons is fixed at 50. Training was performed 10 times for each set of odorants. All weights are quantized to 4 bits. 
    }
    \label{fig:classification result}
\end{figure}

The dataset presented above allows us to estimate the sensor responses for different sets of odorants.
We now proceed further down our processing pipeline by modeling the neural network stage.
To do so, we train a spiking neural network (SNN) to classify odors based on the electrode responses.
The SNN architecture used for this task consists of three fully connected layers (Fig.~\ref{fig:classification result}~left).
The number of input neurons is determined by the number of different ORs used as the sensing front-end, while each output neuron corresponds to a specific odor class. The neurons in the model follow the Leaky-Integrate-and-Fire~(LIF) model, with the subthreshold dynamics described by the following equation:
\begin{equation}\label{eq_lif}
    V_{mem,{i}}(t) = \alpha \, V_{mem,{i}}(t-1) + \sum_{j\neq i} w_{i,j}  \sigma_{j}(t) + \frac{I_{dc}(t) \cdot \Delta t}{C}
\end{equation}
where $V_{mem}$ is the membrane potential of neuron $i$, $j$ is the presynaptic neuron index, $\sigma(t)$ is the spike at time $t$ from neuron $i$ to $j$, $\alpha$ is the membrane decay factor, $w_{i,j}$ denotes the weight of the synapse connecting neuron $i$ to $j$, $\Delta t$ is the time step and $C$ is the membrane capacitor for charge accumulation and leakage. $I_{dc}$ describes the continuous current injected to the membrane from an analog current input. When the membrane potential reaches the threshold value $V_{thr}$ the neuron emits one spike at time $t$.

Training is performed using backpropagation through time (BPTT), an algorithm that involves calculating derivatives across all layers and time steps. Since neuron spikes are discrete, we use a straight-through estimator based on a sigmoid function~\cite{eshraghian2023training}. The SNN architecture provides advantages in processing efficiency during inference due to its massively parallel spiking behavior. Additionally, the implementability of the weight matrix is further optimized through quantization to four bits.

In this case study, the same odor sets (blue, orange, and green) used in the previous subsection were selected for classification~(see Fig.~\ref{fig:odor_space_analysis}). For each odorant, synthetic response traces were generated. The response values for each OR at each time step were organized into a response matrix, where the height of the matrix corresponds to the number of ORs and the width of the matrix represents the number of time steps. This time series was sequentially fed into the multi-layer SNN, column by column, with the analog value passed into the input-to-hidden linear layer, followed by an activation function which defines the behavior of membrane potential update. The following layer operates in the same manner. The spike outputs for each odor category were then summed across all time steps for each sample. The resulting test accuracy is shown in Fig.~\ref{fig:classification result} (right).
As these results demonstrate, the SNN learns to classify the odor sets; however, as expected, the classification accuracy depends on the number of receptors and the chemical structure of the odors.

Note that the SNN classification results do not exactly follow the odor space analysis of Fig.~\ref{fig:odor_space_analysis}; for example, the blue set is classified with lower accuracy, even though it has higher separability.
We attribute this to our definition of separability of odors as the linear distance between odor responses.
Therefore, a non-linear analysis like an SNN can exploit deeper underlying correlations, skewing the classification results across odor sets.
These SNN results can be seen as baseline results, with more complex and biorealistic networks and algorithms required to tackle continual learning, drift compensation, and more complex odor combinations (see Discussion \& Outlook).

\subsection{Electronic sensing pad and neuromorphic hardware implementation of the spiking neural network model}

\begin{figure}[p]
  \centering
  \includegraphics[width=1.0\textwidth]{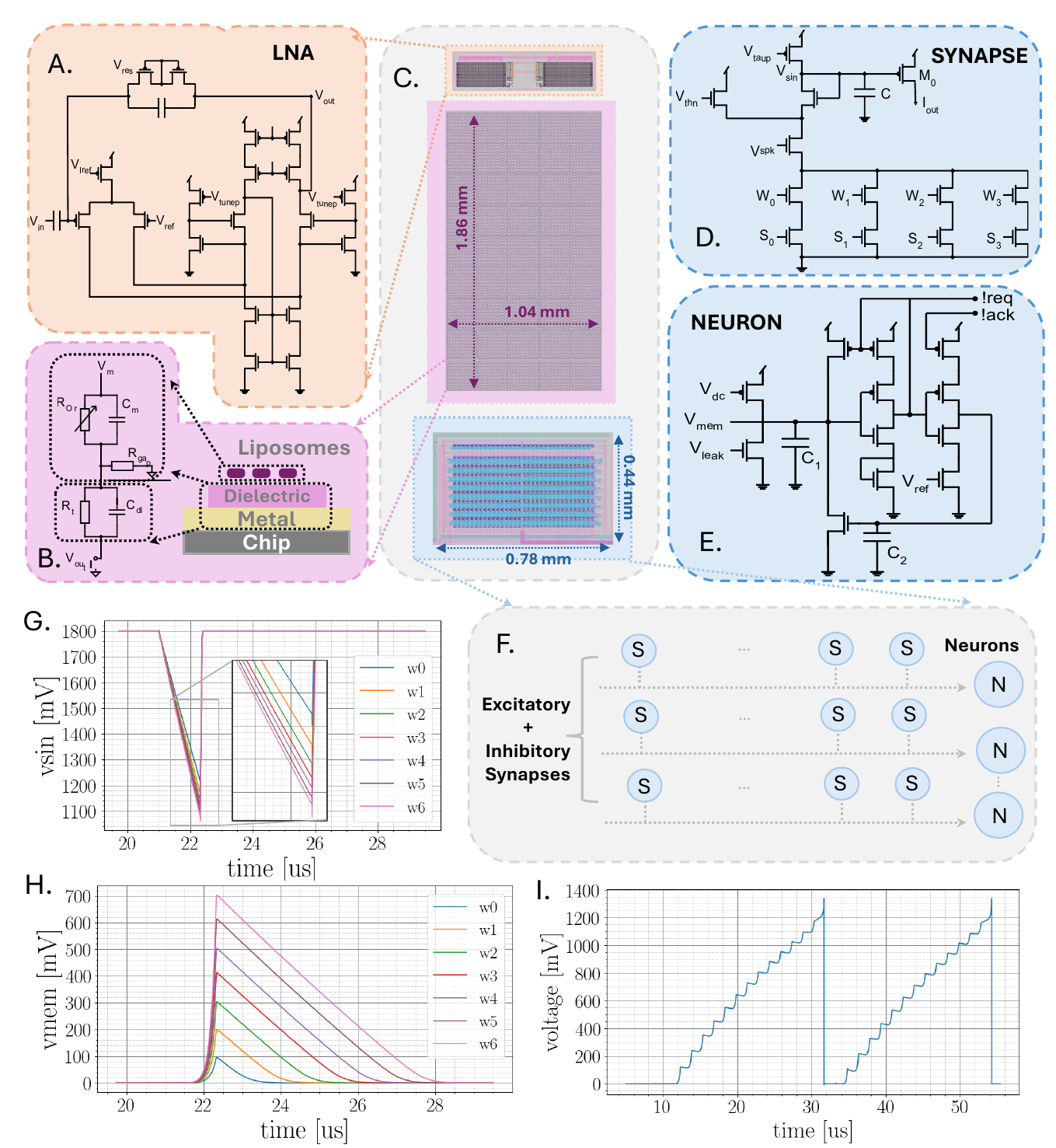}
\caption{Overview of the hardware implementation, including the layout and schematic view of individual blocks. The components are as follows: Low Noise Amplifier (LNA) (A), pads for the sensing front-end and its equivalent circuit schematic (B), and the neuron tile with synapses (D and E). The Ngspice simulation results are shown below: voltage update at node $V_{sin}$ (G), membrane potential across the capacitor $C_{1}$ (equivalence of membrane capacitor) in the NEURON block (E) updates for an equivalent three-bit resolution ranging from 100 mV to 700 mV in seven equal intervals named $w_{0}, .. w_{6}$ (H), and membrane potential integration with low leakage (I). The physical analog circuits support 4-bit binary weight combinations through $S_{0,1,2,3}$ (each switch can be either "1" or "0"), and the binary combinations result in seven distinguishable weight values $w_{0}, .. w_{6}$.}
\label{fig:hardware}
\end{figure}

An illustration of the complete hardware architecture is shown in Fig.~\ref{fig:hardware}. The proposed system consists of an analog front-end designed to process chemosensory signals using an event-driven analog spiking neural network architecture. By bypassing analog-to-digital converters (ADCs) and by leveraging novel synthetic biological assemblies, this approach enables real-time, low-power operation suitable for always-on applications. The system is designed in the SkyWater 130\,nm CMOS process and integrates: i)~a~bioelectro interface that couples lipid vesicles with synthetic olfactory receptors (OR) and microelectrode arrays, and ii)~a~low-noise analog amplifier for signal conditioning. iii)~a~neuromorphic spiking neural network that processes sensor responses in a low-power mixed-signal domain. 

The design exploits transistors operating in the weak inversion (subthreshold) regime. This ensures that tunable device parameters (e.g., bias voltages and currents) can be directly mapped to the computational model, thereby aligning hardware behavior with algorithmic specifications. The range and granularity of these tunable parameters are influenced by physical circuit constraints, including transistor dimensions and the overall network topology. Comparisons between the algorithmically specified neurons and their hardware realizations are critical in verifying the implementation of the design. Therefore, in this section, we simulate the neuron and synapse circuits independently to evaluate their functional behaviors, focusing on how supply and bias voltages affect circuit operation.

Specifically, we provide simulation results of the neuron and synapse circuits separately and analyze their behavior, including voltage dependencies. 

The neuron circuit (Fig.~\ref{fig:hardware}, block E) emulates leaky integrate-and-fire dynamics using the analog charge collected in the capacitor $C_{1}$ to represent the membrane potential $V_{mem}$. Its charging and discharging behavior is controlled by three node voltages: $V_{dc}$ (baseline charging), $V_{leak}$ (leakage or decay), and $V_{ref}$ (refractory reset). Input spikes from previous layers arrive through an analog differential-pair synapse circuit (Fig.~\ref{fig:hardware}, block D)~\cite{bartolozzi2007synaptic}. Here, digital weights $W[0\sim3]$ determine the strength of each bit, and switches $S[0\sim3]$ enable their contribution. The combined synaptic current is injected into the neuron's membrane node, shaping its temporal dynamics (Fig.~\ref{fig:hardware} I). Spiking events are communicated asynchronously to a microcontroller via a common neuromorphic address event representation scheme that uses a 4-phase handshake protocol. This means exploiting digital $!req$ and $!ack$ signals to ensure reliable transfer of each spike without a global clock.
We refer the reader to the Methods section for more detailed circuit analysis.

\newpage
\subsection{Hybrid synthetic sensory neuron}
\begin{figure}[p]
    \centering
    \includegraphics[width=1.0\textwidth]{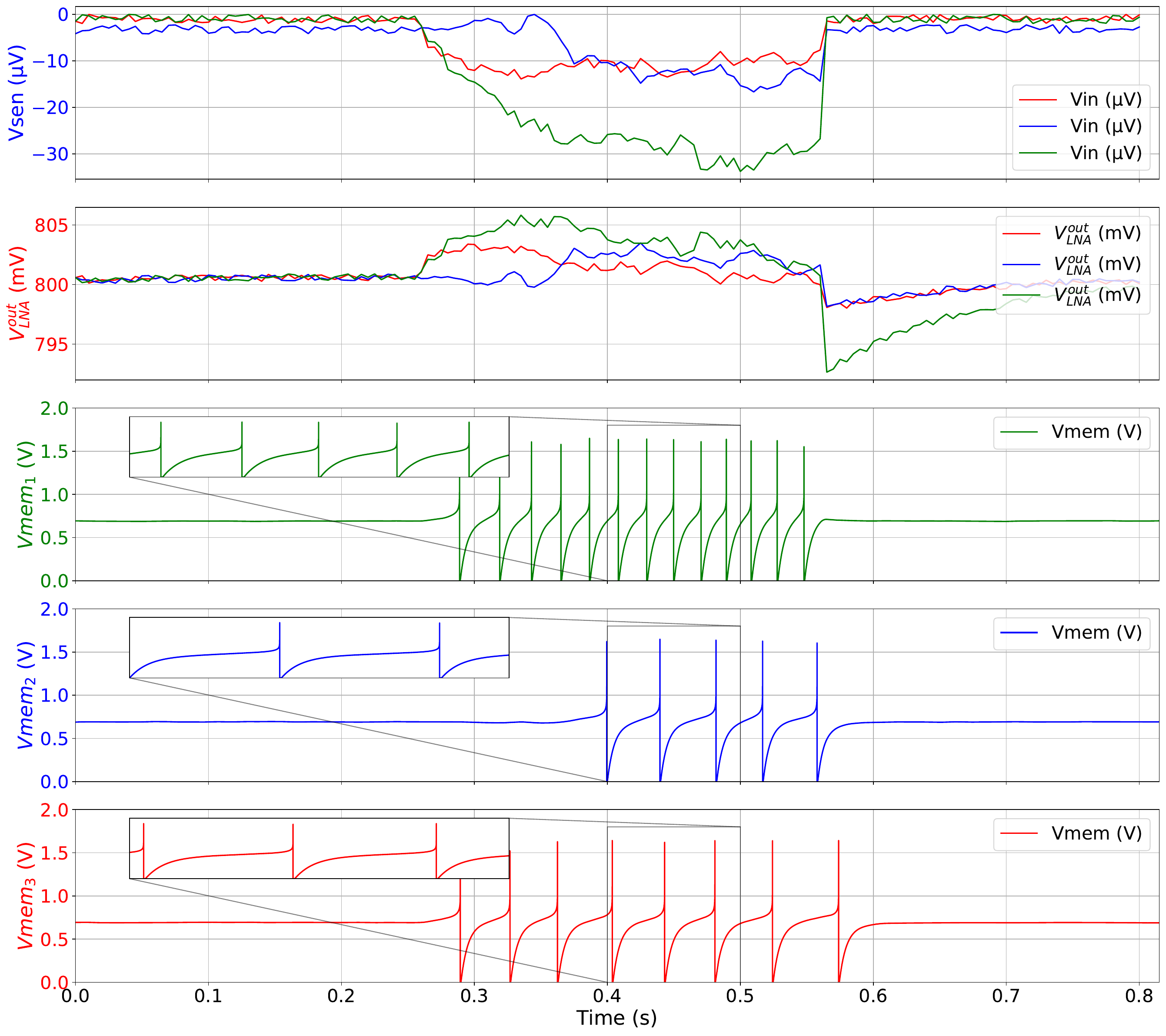}
    \caption{Ngspice simulation results for a hybrid synthetic sensory neuron for analog-to-spike coding: The top plot illustrates three input traces derived from the synthetic dataset, which are applied to the LNA's input. The second plot displays the LNA's voltage output, demonstrating an output swing of approximately 100 mV with a gain of 50.05 dB. This amplified output is then utilized to modulate three current injection nodes ($V_{dc}$), which directly stimulate the somatic circuits, triggering neuronal firing as shown in the last three plots.}
    \label{fig:lna-neuron-encoding}
\end{figure}

We  simulated the full signal processing pipeline from the ion dynamics on the microelectrode to the firing of VLSI spiking neurons. The microelectrode pad array is responsible for detecting low-amplitude bioelectronic signals in the tens of microvolt range (as shown in Fig.~\ref{fig:lna-neuron-encoding} top), which are then fed into a low-noise amplifier (LNA) for signal conditioning. The LNA enhances weak signals, providing an output swing of approximately 5-10 mV with a gain of 50.05 dB, ensuring reliable signal amplification while preserving fidelity (second plot in Fig.~\ref{fig:lna-neuron-encoding}). This amplified signal is then processed by spiking neurons, where the modulated activity of current injection is provided by connecting the LNA output, after scaling, to the ($V_{dc}$) nodes, directly stimulating somatic circuits to generate spikes. The event-driven nature of this architecture enables efficient signal encoding, where spike generation is proportional to the detected signal strength, facilitating mean rate coding for robust event-based detection (three bottom plots in Fig.~\ref{fig:lna-neuron-encoding}). A comprehensive simulation captures the entire signal processing pipeline, where soma circuits are maintained near the threshold by balancing the contributions of injection and leakage currents (\(V_{dc}\) and \(V_{leak}\)), allowing neurons to rapidly generate spikes upon detection of simulated odor signals on the pad. 

By integrating synthetic biological components with VLSI circuits in this manner, a new class of neuron emerges, the \textit{hybrid synthetic sensory neuron}, wherein the hybrid interface from pad to amplifier to soma enables event-driven detection and spike generation. 

\FloatBarrier

\section{Discussion \& Outlook}

A key component of our platform is the bioelectronic interface, which bridges biological odor recognition with CMOS-based electronics. To enhance signal transduction, it will be necessary to characterize the synthetic cell-electrode interface and  optimize the electrode impedance for improved sensitivity and signal integrity. 
We are relatively confident to be able to obtain a high electrical signal fidelity, as systematic engineering of electrodes for biosignal detection is well established, for example, by polymer layers \cite{wang2021impedance}. A larger conceptual problem is the limitation to sensing of biocompatible molecules, e.g. avoiding substances that might disintegrate the synthetic cells, and in general the limitation in the lifetime of biological receptors outside of living systems. After all, there will be good reasons why living cells invest in turnover (disintegration and synthesis) of membrane proteins with typical half-lives of days \cite{dorrbaum2018local}. We therefore believe that disposable sensor elements are a realistic first implementation of our design. Due to the low power consumption of neuromorphic systems, enabling compact design, many uses of disposable sensors come to mind in the domains of environmental sampling, safety, and food processing~\cite{cheng2021development}.
Moreover, by integrating several chips into the platform, each microelectrode operating in its own bath and targeting a single OR type, the system could be scaled into a multi-microelectrode architecture. This arrangement would allow simultaneous screening of multiple ORs while still allowing incremental testing as needed.

In addition, while process variations and mismatch in weak-inversion circuits introduce variability in neuromorphic systems~\cite{benjamin2023analytical} our system integrates multiple biasing mechanisms to compensate for these effects, which will be further validated during hardware testing. However, it is worth noting that bio-inspired computational strategies often exploit the variability in all the network parameters as a computational resource  for example in learning applications~\cite{thakur2017analogue}, or in dynamical computation using attractor networks \cite{camilleri2010self} and decision perceptual making~\cite{corradi2015decision} tasks; we plan to test these computational primitives in the lab with real chemosensory data. 
In our system, we expect the bigger source of drift and variability to stem from the biological receptors, for example by leakage of ions from the synthetic cell vesicles which has a typical timescale of days ~\cite{verkman1989quantitative}. This could make it necessary to adapt electrical methods for compensation, e.g.~DC drift offset compensation, or more systematic approaches like homeostatic learning mechanisms~\cite{cannon2016synaptic}. 

While the actual hardware implementation of hybrid synethtic sensory neurons is underway,
we plan to extend the current work with hardware characterization, comprehensive dataset collection, and demonstration of system scalability. Specifically, efforts will be directed toward expanding the platform to support multiple sensor pads, each featuring distinct odor receptor chambers, thereby enhancing the system's capability to distinguish between a wider range of chemical compounds, and our mixed-signal spiking neural network hardware classifiers will be used to achieve real-time classification of complex odor mixtures.

One important stipulation of our simulation pipeline are the underlying odorant concentrations levels.
To model odor responses, we have used the DoOR dataset, which contains recordings with high concentrations of volatile compounds, orders of magnitude higher than naturally occuring concentrations.
While this increases the signal-to-noise ratio in spike recordings, it also reduces the selectivity of individual ORs~\cite{wachowiak2025recalibrating,dennler2025neuromorphic}.
I.e.,~there is a trade-off between sensitivity and selectivity in neuron responses, and we expect this to be present also in our hybrid synthetic sensory neurons.
In a real-world application of our sensing platform with natural concentrations, this trade-off is likely to manifest as a lower signal-to-noise ratio, but tighter matching of individual ORs to odors.
It is therefore of high interest to repeat the odor space analysis under natural and artifically high concentrations using our sensing system.
This will not only help us fine-tune the performance of our platform,
but may also provide insights about the sensitivity/selectivity curve of sensory neurons in vivo.

In addition, as part of our co-design, we are developing network models which are able to extract relevant information delivered from the sensors, such that high classification accuracy with minimal number of ORs can be achieved. 
A crucial data feature may be the temporal structure of the stimulus~\cite{olsson2006chemosensory}.
The hardware simulations shown here are based on the direct injection of the sensor current into the first layer of neurons, effectively resulting in rate coding with a bin width of~$\mathcal{O}(200)\,$ms~(Fig.~\ref{fig:lna-neuron-encoding}).
By adapting the bias currents that control the parameters of the synapses and neurons circuits, such as time constants, leak currents, etc., we aim to fully exploit the temporal structure of the stimulus.
In parallel, coding schemes such as precise spike timing have been observed in our model species~\cite{egea2018high}, and may prove valuable for our application.

The setup described here is limited to supervised training using single (non-mixed) odor presentations.
Future work will investigate odor mixtures as they appear in real-world applications.
In this regard, it may prove valuable to model ligand-OR interactions as fuzzy sensors, as proposed in~\cite{gentili2014human}.
Due to OR co-activation, it is unclear which response to expect from the synthetic cells under odor mixtures, and careful characterization will be needed.
Finally, we aim to investigate unsupervised and few-shot learning for real-world use of our sensing platform, following recent progress in understanding continual learning in the fruit fly olfactory system~\cite{10.1162/neco_a_01615}.

Moving beyond insect olfaction, recent advances have also demonstrated successful integration of human olfactory receptors (hORs) with organic neuromorphic synaptic devices, achieving highly accurate, molecule-specific odor discrimination~\cite{song2024pattern}. Unlike our approach, which employs synthetic insect olfactory receptor complexes coupled to analog CMOS spiking networks, Song et al.~\cite{song2024pattern} used part of the human olfactory system interfaced with organic electronics, showing precise discrimination of structurally similar odorants. Although their work highlights the feasibility of exploiting human olfactory specificity for targeted sensing applications, our framework emphasizes modularity and adaptability through synthetic biology and neuromorphic codesign, aiming to  address broader olfactory detection challenges.
Finally, Wu et al.~\cite{wu2025bionic} proposed a bionic olfactory neuron by combining organic field-effect transistor arrays with in-sensor reservoir computing to achieve classification for complex odor mixtures, isomers, and homologs, however, our methodology is complementary as we uniquely integrates synthetic biology-based receptors with neuromorphic circuits, potentially enabling direct biological specificity and real-time adaptability.

\section{Conclusion}

This work presents a novel co-design methodology of a neuromorphic chemosensing system that aims at integrating biological olfactory receptors with analog neuromorphic circuits, mimicking natural olfactory processing. We demonstrate, using computer simulations, the use deep-learning tools such as AlphaFold to select suitable odorant receptors (ORs) and we generate a synthetic dataset in which each Orco-OR complex is modelled as an independent two-state Markov process. The resulting open/close transitions yield continuous ion-channel currents that our electrode model converts into analogue voltage traces (Fig.~\ref{fig:synth_dataset_traces}). These noisy, microvolt-level signals feed directly into an NGSpice simulation of the neuromorphic front-end (Fig.~\ref{fig:lna-neuron-encoding}), where sub-threshold analogue neurons transform them into spikes. A spiking neural network, supervisely trained on the same odors, then classifies the spike patterns with high accuracy (Fig.~\ref{fig:classification result}). Importantly, by combining biological odor receptors with neuromorphic hardware, we aim to achieve real-time, energy-efficient chemosensory detection. The use of weak-inversion transistors and event-driven spiking neural networks ensures on-demand processing, minimizing power consumption while maintaining high sensitivity to chemical stimuli. 

Our approach is unique in that it combines the co-design of the physical realization of sensory neurons, processing, and learning mechanisms, all in one device. To reach this goal, we build on recent advances in the use of machine learning methods in biology to enable the co-design with established electrical engineering methods. 

The results of our model pipeline give us strong guardrails for the development of our platform,
informing the co-design of Orco-OR complexes, vesicle properties, amplification stages, and decoding hardware.
As seen in Fig.~\ref{fig:classification result}, the classification of odors is confounded by the noise and the number / types of available OR.
In this regard, we interpret the SNN classification results as a baseline.

\section{Materials \& Methods}

\subsection{Orco-OR complex stability}
\label{subsec:oroc_pulldown}
Odor receptor amino acid sequences were obtained from UniProt [92], with the search query '(protein\_name:``odorant receptor'') AND (taxonomy\_id:XXXX)'. Species associated with olfactory research were selected (see Supplementary Material for used taxonomy ids).
The list was manually cleaned for duplicates and fragment sequences. The total of 348 sequences were assembled into 1:1 complexes with the Drosophila melanogaster Orco channel (Uniprot id Q9VNB5) using Alphapulldown scripts [93]. Complex structures were predicted using AlphaFold v2.3.1. In addition to generation of~.pdb~files that were further analyzed with Chimera (ChimeraX 1.9), the script produced a list of the complexes with different scoring metrics. Two of the scores were used to filter for accurate and stable complexes.
The dataset  (252 insect species, 48 Dm and 48 \textit{M.~musculus} complexes with Dm Orco) was first sorted using the ipTM score applying a cut-off of~0.6. The ipTM score is derived from pairwise distance error predictions for interface residues \cite{evans2021protein, jumper2021highly}. For complexes above ipTM > 0.6  the interface solvation energy was calculated from the residues contributing to the protein-protein interface \cite{kraml2019solvation}.  

\subsection{Molecular Docking}
\label{subsec:docking}
AlphaFold (version 2022-11-01) structures of olfactory receptors were obtained from Uniprot. Potential binding sites were identified using P2Rank~\cite{krivak2018p2rank}. Terpenes were represented using their respective SMILES strings. The SMILES structures were converted into 3D coordinates using RDKit (version 2024.9.5). Docking was performed using AutoDock Vina \cite{trott2010autodock} with an exhaustiveness parameter set to 32. The search space was defined based on the P2Rank-predicted binding pocket coordinates in a box of 5×5×10~Å. The binding energy of the most favorable pose was reported.

\subsection{Electrode model}
\label{subsec:methods_electrode}
Multielectrode Arrays (MEAs) consisting of 60 titanium nitride electrodes of 30~$\mu$m diameter (Model 60tMEA200/30iR-ITO-gr, Multichannel Systems, Germany) were characterized using impedance spectroscopy. Impedance spectra in the frequency range 10 kHz to 1 MHz were obtained in phosphate buffer solution of physiological salinity at room temperature electrically connected by a silver-chloride ground electrode.
A three-circuit model consisting of a resistor in series with capacitance~($C_{dl}$) and a resistor ($R_t$) in parallel was used to fit the spectra with $\chi^2 \approx 0.2 $. The resistance of the gap $R_{gap}$ was calculated from $R_{gap} = \rho_s /(5 \pi d_{gap}$), with approximate values for the solution of conductivity $\rho_s = 1$ $\Omega$m and $d_{gap} = 1$ nm \cite{STEINKUHLER20161454,PhysRevE.55.877}.
The final values for the electrode model are $\Rgap = 6.37 \times 10^{7} \, \Omega$, $\Cdl = 8.99 \times 10^{-11} \, \text{F}$, and $\Rt = 7.92 \times 10^8 \, \Omega$.

\subsection{Odor space analysis}
\label{subsec:methods_odor_space_analysis}

We describe our method for selecting the most viable \acs{OR}s for a given set of odorants in Algorithm~\ref{alg:OR_reduction_algo}.
The method requires a response data matrix with entries (odor $\times$ \acs{OR}), which we have retrieved from the DoOR~2.0.1 database~\cite{Galizia2010,munch_door_2016};
specifically, we use the provided $\texttt{door\_response\_matrix.csv}$, which contains responses of 78 sensory unit types to 692 odorants, normalized to $[0,1]$ across all responding units.
We first exclude all unit types not clearly linked to a single \acs{OR} type, as described in \cite{munch_door_2016}, which leaves a total of 52~\acs{OR}s to be analyzed.
As only a subset of all unit type/odor combinations have been measured (about 14\,\%), we replace missing entries~($\texttt{NA}$) with zero response.
This may introduce the issue that some odors are separated only by the absence of any response;
in practice, this would make it impossible to distinguish such an odor from noise currents when no ligand is present.
We mitigate this issue by including a `no ligand/SFR (spontaneous firing rate)' category in the analysis, see below.

From the response data, we select only the odorants (rows) in which we are interested.
Our method then iteratively removes \acs{OR}s (columns) from the response matrix, until the desired number $\NOR$ of \acs{OR}s to be reconstituted is reached.
The \acs{OR} to be removed is selected by the criterion of `odor separability', which is the sum of all Euclidean distances of rows (odorants).
This means that we exclude one \acs{OR} from the data, and compare the response of the remaining \acs{OR}s to the odors.
The higher the Euclidean distance, the higher the odor separability.
We find the least important \acs{OR} by comparing the separabilities with that \acs{OR} excluded: If we denote as $D_{\setminus j}$ the responses of all \acs{OR}s minus receptor $j$, then that $D_{\setminus j}$ which has the highest separability is the one we want to continue with.
We thus pick the receptor $j$ for which the separability of $D_{\setminus j}$ is reduced the least, and eliminate it from the pool of receptors.
This procedure is repeated iteratively until the desired number $\NOR$ is reached.

Our algorithm for selecting odor/\acs{OR} sets is easily implemented and performs the reduction quickly.
Note however that the underlying dataset corresponds to response measurements of olfactory receptor neurons corrected for the baseline firing rate, and doesn't cover cases where no ligand is present at all, as described above.
Therefore, if the final selection given by our algorithm includes odors for which none of the selected \acs{OR}s respond at all, this odor will be experimentally indistinguishable from no ligand binding.
To cover this case, we have included a `no ligand' category, i.e.~a row in~$D_\text{all}$ which corresponds to the spontaneous firing rate.

\begin{algorithm}[H]
\caption{OR reduction algorithm
}\label{alg:OR_reduction_algo}
\begin{algorithmic}
\Require desired number of \acs{OR}s $\NOR$, indices of desired odorants $\{i\}$,
response~data~$D_\text{all}$ with entries (odor $\times$ OR)\\
\State $N \gets \NOR^\text{total}$   \Comment{total number of columns in dataset $D_\text{all}$}
\State $D \gets D_\text{all}[\{\text{rows }i\}]$    \Comment{only take odorants we are interested in}\\
\While{$n > \NOR$}      \Comment{until desired number of ORs is reached}
\For{$j$ in columns of $D$}         \Comment{for each OR $j$}
    \State $D_{\setminus j} \gets D[\{\text{columns } \neq j\}]$      \Comment{exclude OR $j$ responses}
    \State $S(D_{\setminus j}) \gets \sum_{m\neq n} \|$ row $m$ of $D_{\setminus j}\, -$ row $n$ of $D_{\setminus j}$ $\|$                      \Comment{calc.~odor~separability}
\EndFor
\State $j_\text{elim} \gets \text{argmax}_j \big[ S(D_{\setminus j}) \big]$\Comment{select OR for which~highest separ.~remains}
\State $D \gets D[\{\text{columns} \neq j_\text{elim}\}]$ \Comment{eliminate that OR}
\State $n \gets n - 1$
\EndWhile
\end{algorithmic}
\end{algorithm}

\subsection{Synthetic dataset}
\label{subsec:methods_synth_dataset}

The synthetic dataset is modeled using a Markov model of Orco ion channels.
We assume that each synthetic cell only has one specific OR type, and that there are $\NORCO$ ion channels per synthetic cell in proximity to each electrode.
In order to model Orco realistically, we compare our simulation with the electrophysiological characterization of VUAA1-induced Orco currents, matching the data of Extended Data Fig.~2 in Butterwick et al.~2018~\cite{Butterwick2018}.

To generate a voltage trace of one Orco for a given odor/OR type combination, we perform the following steps:
\begin{enumerate}
\item fetch the normalized odor/OR response $d_\text{odor,OR}$ from the response dataset $D$.
\item simulate a two-state Markov model with channel opening and closing probabilities $\popen$ and $\pclose$ for $(\Tdata \times \dt)$ steps (`data trace').
The channel opening probability is given by $\popen = \pclose \times d_\text{odor,OR}$.
\item convert the open and closed states to currents in pA by drawing from $\mathcal{N}(\mu_\text{open}, \sigma_\text{open})$ and $\mathcal{N}(\mu_\text{closed}, \sigma_\text{closed})$, respectively.
\item generate ion channel noise for $(\Ttotal \times \dt)$ steps and filter to match the noise found in~\cite{Butterwick2018}.
To match the magnitude and frequency spectrum, we compare to the parts of the recording where no ligand is present.
\item randomly shift the data trace in time and add to the noise trace.
\end{enumerate}
For each synthetic cell, we perform these steps $\NORCO$ times and sum the current traces.
Finally, we obtain the voltage induced in the electrode by modeling an RC circuit which integrates the generated current.
The parameters $\Rgap$, $\Cdl$ and $\Rt$ are determined from impedance spectroscopy (see Section~\ref{subsec:methods_electrode}).
The transition probabilities $\{\popen, \pclose\}$ of the Markov model were estimated based on Orco measurements from Ref.~\cite{Butterwick2018}.
In particular, these recordings show baseline noise~$\mathcal{O}(0.5\, \text{pA})$, some signal currents of about $2 \, \text{pA}$, and larger portions of signal showing current flow of 2 to $5 \, \text{pA}$.
The former can be attributed to a single Orco opening, while the latter is potentially caused by multiple Orcos being probed at the same time.
In order to make a conservative estimate of a single Orco current, we base the Markov model parameters only on a portion of the signal which contains current flow up to $2 \, \text{pA}$ (central region in Extended Data Fig.~2c of \cite{Butterwick2018}, corresponding to the range from 48.9\,s to 49.6\,s in the recordings provided by the authors).
Similarly, the parameters $\{\mu_\text{open}, \sigma_\text{open}\}$ and $\{ \mu_\text{closed}, \sigma_\text{closed} \}$ have been chosen to match the binned data generated from the same region of measurement.

For all parameters and implementation details, see the code repository~\cite{SYNCH_code_repository}.

\subsection{Design and Simulation of the Neuromorphic Hardware Platform}

\begin{figure}[htbp]
    \centering
    \includegraphics[width=0.5\textwidth]{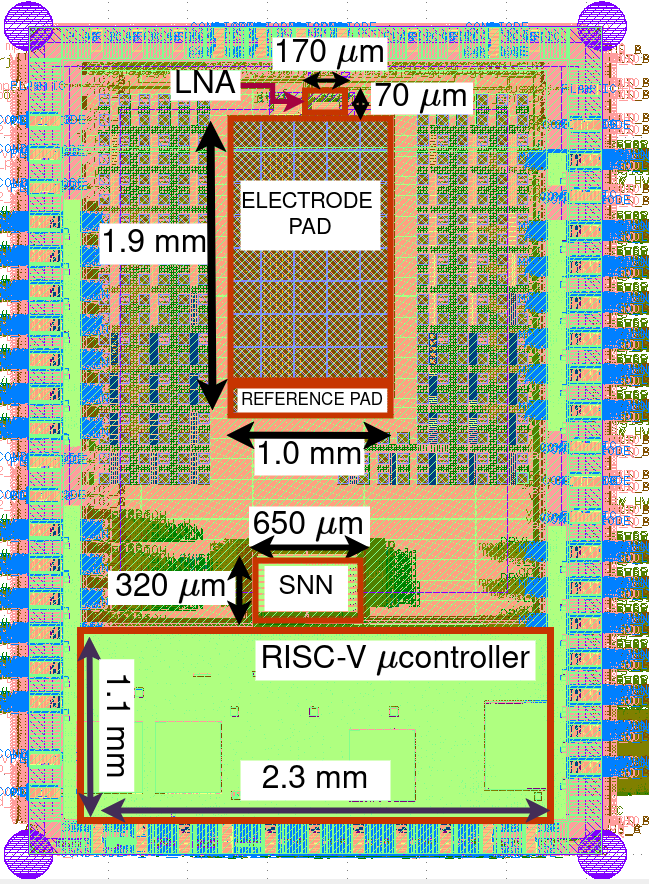}
    \caption{The full chip layout in SkyWater 130 nm features an electrode pad interface (1.6×1.0~mm), a reference pad (0.2×1.0~mm), low-noise amplifier (170×70~$\mu$m), a mixed-signal spiking neural network (SNN) processor (650×320~$\mu$m), and a 32-bit RISC-V (RISCV32IMC) $\mu$controller (1.1×2.3 mm). }
    \label{fig:hardware_layout}
\end{figure}

The system integrates an analog front-end with an event-driven, asynchronous spiking neural network and a microcontroller. It consists of: (i) a bioelectronic interface coupling lipid vesicles with synthetic olfactory receptors and microelectrode arrays, (ii) a low-noise analog amplifier for signal conditioning, (iii) a spiking neural network for energy-efficient sensor response processing, and (iv) a RISC-V microcontroller for spike decoding and synaptic weight configuration. Fig.~\ref{fig:hardware_layout} illustrates the layout of the full chip. 
The low-noise amplifier (LNA) exploits a folded cascode OTA (FC-OTA) with an active cascode and a flipped current follower (FCF) to improve signal integrity, transconductance efficiency, and noise performance. The FC-OTA topology enhances gain while maintaining a wide output swing, the active cascode increases output impedance for better gain stability, and the FCF reduces impedance at the folded node, ensuring efficient current transfer. A detailed overview of these techniques is provided, with further details in the work by R. Sanjay et al.~\cite{sanjay2018low}.
Somatic and synaptic circuits are designed with weak-inversion transistors and follow the principles of well-known neuromorphic designs~\cite{bartolozzi2007synaptic,indiveri2011neuromorphic} that have been demonstrated to scale to advanced technologies nodes~\cite{qiao2017analog, eslahi2022compact}.

The schematic of the neuron circuit is illustrated in Fig.~\ref{fig:hardware}, block E. It uses a capacitor $C_{1}$ to represent the membrane potential $V_{mem}$. Three analog control voltages regulate the behavior of $V_{mem}$: 
\begin{itemize}
    \item $V_{dc}$ sets the baseline charging current, 
    \item $V_{leak}$ defines the rate of decay, and 
    \item $V_{ref}$ determines the reset level during the refractory period.
\end{itemize}
This arrangement enables the hardware circuit to emulate a leaky integrate-and-fire (LIF) neuron. When $V_{mem}$ is below the firing threshold, its dynamics follow:
\begin{equation}\label{eq_lif_dynamics}
    V_{mem}(t+\Delta t) = \alpha V_{mem}(t) + I_{dc} \cdot \Delta t
\end{equation}
where $\alpha$ models the leakage over time. Unlike idealized models, here $\alpha$ depends on the membrane potential itself due to the circuit’s analog characteristics:
\begin{equation}
    \alpha = 1 - \frac{R_{leak} \cdot \Delta t}{V_{mem}}
\end{equation}
with $R_{leak}$ as the leakage rate (V/s), and $\Delta t$ as the simulation time step.

The neuron operates in an event-driven (asynchronous) manner, emitting spikes whenever $V_{mem}$ crosses the firing threshold. To communicate these events reliably to a digital system like a microcontroller, we implement a 4-phase handshake protocol using two signals: $!req$ (request) and $!ack$ (acknowledge).

\begin{enumerate}
    \item When a spike occurs, the neuron raises $!req$ to signal an event.
    \item The receiver detects this and responds by asserting $!ack$.
    \item Upon receiving acknowledgment, the neuron deasserts $!req$.
    \item Finally, the receiver lowers $!ack$, completing the handshake.
\end{enumerate}

This protocol ensures accurate and lossless communication of spikes, even in systems without a global clock.

Fig.~\ref{fig:hardware}, block D shows the synapse circuit. Synaptic inputs are implemented using a 4-bit digital representation. Each weight bit $W[i]$ (where $i = 0\ldots3$) defines the current strength, and its corresponding switch $S[i]$ enables or disables that contribution. When active, the current from each bit is summed to produce a total synaptic output current $i_{out}$:
\begin{equation}
    i_{out} = \sum_{i=0}^{3} S[i] \cdot W[i]
\end{equation}
This current is injected directly into the neuron’s $V_{mem}$ node, modifying its potential.

To balance the influence of bias (from $V_{dc}$) and synaptic input, transistor sizing in these branches is carefully chosen. The synaptic current is derived from subthreshold MOSFET operation, yielding an exponential current-voltage relationship. The discharge rate of the intermediate node $V_{sin}$ follows:
\begin{equation}\label{eq_wi_MOS}
    R_{V_{sin}} = \frac{dV_{sin}}{dt} = \frac{I_{tot}}{C} = \frac{1}{C} \sum_{i} \frac{I_{ds,0_i} \cdot W_{i}}{L_{i}} e^{\frac{V_{W_{i}} - V_{th,n}}{nU_{T}}}
\end{equation}
Here, $V_{th,n}$ is the threshold voltage, $U_T = \frac{kT}{q} \approx 25$~mV at 300~K, and $n = 1 + \frac{C_{D}}{C_{ox}}$ is the subthreshold slope factor. The final synaptic current $i_{out}$ is produced via a P-type MOSFET $M_0$ and injected into the membrane node.

The simulation results of the hardware platform are obtained using Ngspice. The complete hardware system is currently being manufactured by Efabless and shown in Figures~\ref{fig:hardware} and \ref{fig:hardware_layout}, where we provide a detailed overview of the device, illustrating the four key components and the chip layout. 
The reference pad serves as a stable biasing point for the chip, allowing the analog front-end and neural processing circuits to operate with respect to a well-defined potential. This is especially important for minimizing common-mode noise and ensuring reliable signal transduction from the bioelectronic interface.
The spiking neural network includes a total of 16 neurons and 512 synapses, organized to support configurable connectivity patterns. This configuration enables efficient encoding of sensory inputs and supports spike-driven odor classification.

\FloatBarrier

\section*{Acknowledgments}

This work is part of the project \textit{SYNCH: Combining SYnthetic Biology \& Neuromorphic Computing for CHemosensory perception}, funded by the Volkswagen Foundation under the call “NEXT –- Neuromorphic Computing”.
We gratefully acknowledge the foundation's initiative to unite diverse disciplines in the pursuit of novel scientific ideas and their continued support in making this research possible.
We also thank Josefina del M{\'a}rmol for providing recorded data characterizing Orco and helpful explanations.

    \printbibliography

\appendix

\end{document}